\documentclass[conference]{IEEEtran}
\usepackage{times}

\usepackage[numbers]{natbib}
\usepackage{multicol}
\usepackage[bookmarks=true]{hyperref}
\usepackage{amsmath}
\usepackage{amssymb}
\usepackage{amsfonts}
\usepackage{subcaption}
\usepackage{graphicx}
\usepackage{array}
\newcolumntype{L}[1]{>{\centering\arraybackslash}m{#1}}
\newcolumntype{R}[1]{>{\hsize=#1\hsize\raggedleft\arraybackslash}X}
\newcolumntype{C}[2]{>{\hsize=#1\hsize\columncolor{#2}\centering\arraybackslash}X}
\usepackage{threeparttable}
\usepackage{multirow}
\usepackage{booktabs}

\begin{document}

\title{An Empowerment-based Solution to Robotic Manipulation Tasks with Sparse Rewards}

% You will get a Paper-ID when submitting a pdf file to the conference system
% \author{Author Names Omitted for Anonymous Review. Paper-ID [6]}

\author{\authorblockN{Siyu Dai\authorrefmark{1},
Wei Xu\authorrefmark{2},
Andreas Hofmann\authorrefmark{1} and
Brian Williams\authorrefmark{1}}
\authorblockA{\authorrefmark{1}Computer Science and Artificial Intelligence Laboratory, Massachusetts Institute of Technology}
\authorblockA{\authorrefmark{2}Horizon Robotics\\
Corresponding email: sylviad@mit.edu. Work partially done during an internship at Horizon Robotics}
}

% avoiding spaces at the end of the author lines is not a problem with
% conference papers because we don't use \thanks or \IEEEmembership

%Homer Simpson\authorrefmark{2},
%Montgomery Scott\authorrefmark{3} and
%Eldon Tyrell\authorrefmark{4}}
%Telephone: (800) 555--1212, Fax: (888) 555--1212}
%\authorblockA{\authorrefmark{4}Tyrell Inc., 123 Replicant Street, Los Angeles, California 90210--4321}}

\maketitle

\begin{abstract}
In order to provide adaptive and user-friendly solutions to robotic manipulation, it is important that the agent can learn to accomplish tasks even if they are only provided with very sparse instruction signals. 
To address the issues reinforcement learning algorithms face when task rewards are sparse, this paper proposes an intrinsic motivation approach that can be easily integrated into any standard reinforcement learning algorithm and can allow robotic manipulators to learn useful manipulation skills with only sparse extrinsic rewards. Through integrating and balancing empowerment and curiosity, this approach shows superior performance compared to other state-of-the-art intrinsic exploration approaches during extensive empirical testing. Qualitative analysis also shows that when combined with diversity-driven intrinsic motivations, this approach can help manipulators learn a set of diverse skills which could potentially be applied to other more complicated manipulation tasks and accelerate their learning process.
\end{abstract}

\IEEEpeerreviewmaketitle

\section{Introduction}

Real-world robotic manipulation tasks are diverse yet often complicated. An ideal robotic agent should be able to adapt to new environments and learn new tasks by exploring on its own, instead of requiring intensive human supervision. The traditional task and motion planning approach to robotic manipulation~\citep{kaelbling2013integrated} typically requires a significant amount of domain-specific prior knowledge, and acquiring this knowledge often involves intensive human engineering. On the other hand, reinforcement learning (RL) agents have demonstrated impressive performances in scenarios with well-structured environment and dense reward signals~\citep{levine2016end}. However, learning-based approaches to manipulation typically only work well when the reward function is dense or when expert demonstrations are available. This is because when the state and action space is high-dimensional and the reward signal is sparse, RL agents could potentially spend a long time exploring the state space without getting any reward signal. Therefore, RL has seen less success in tasks with unstructured environments like robotic manipulation where the dynamics and task rewards are less intuitive to model. 

Designing task-specific dense reward functions to simplify the sparse-reward RL problem has been a common solution for manipulation problems, but in most practical applications, hand designing dense reward functions for every robot in every task and every environment is infeasible and might bias the agent's behavior in a suboptimal way~\citep{andrychowicz2017hindsight}. Inverse reinforcement learning approaches seek to automate reward definition by learning a reward function from expert demonstrations, but inevitably demand a significant amount of task-specific knowledge and place considerable expert data collection burden on the user~\citep{riedmiller2018learning}. Recent advances in meta-learning allow agents to transfer learned skills to other similar tasks~\cite{finn2017model, chitnis2019learning}, but a large amount of prior meta-training data across a diverse set of tasks is required, which also becomes a burden if a lot of human intervention is needed. Therefore, effectively solving sparse reward problems from scratch is an important capability that will allow RL agents to be applied in practical robotic manipulation tasks. 

In this paper, we propose an empowerment-based intrinsic exploration approach that allows robots to learn manipulation skills with only sparse extrinsic rewards from the environment.
% Describe the approach and the main innovative points; mention (and cite) the simulation environment
% However, how to solve sparse reward problems from scratch without human supervision still remains an open problem despite many trials in the reinforcement learning field.
Empowerment is an information-theoretic concept proposed in an attempt to find local and universal utility functions which help individuals survive in evolution by smoothening the fitness landscape~\citep{klyubin2005empowerment}.
Through measuring the mutual dependence between actions and states, empowerment indicates how confident the agent is about the effect of its actions in the environment. In contrast to novelty-driven intrinsic motivations which encourage the agent to explore actions with unknown effects, empowerment emphasizes the agent's ``controllability'' over the environment and favors actions with predictable consequences. We hypothesize that empowerment is a more suitable form of intrinsic motivation for robotic manipulation tasks where the desired interactions with environment objects are typically predictable and principled. Imagine a robot interacting with a box on the table. Intuitively, the undesirable behaviors of knocking the box onto the floor should generate higher novelty since it helps explore more states that haven't been visited, and the desirable behaviors of pushing the box or lifting the box up should generate higher empowerment because the effects of these actions are more predictable. 

Based on this intuition, we apply an empowerment-based intrinsic motivation to manipulation tasks with sparse extrinsic rewards and demonstrate that with the help of novelty-driven rewards at the beginning of training, neural function approximators can provide reasonable estimations of empowerment values. With extensive empirical testing on object-lifting and pick-and-place tasks in simulation environments, we show that this empowerment-based approach outperforms other state-of-the-art intrinsic exploration methods when the extrinsic task rewards are sparse. Although the concept of empowerment has previously been discussed in the context of RL~\citep{mohamed2015variational}, to the author's best knowledge, this paper is the first successful demonstration of the effectiveness of empowerment in terms of assisting RL agents in learning complicated robotics tasks with sparse rewards.

\section{Related Work}

Reinforcement learning for sparse reward tasks has been been extensively studied from many different perspectives. Curriculum learning~\cite{bengio2009curriculum} is a continuation method that starts training with easier tasks and gradually increases task difficulty in order to accelerate the learning progress. 
% It has seen success in many applications including language modeling~\cite{graves2017automated}, autonomous navigation~\cite{mirowski2018learning, ivanovic2019barc} and robotic manipulation~\cite{florensa2017reverse}. 
However, many curriculum-based methods only involve a small and discrete set of manually generated task sequences as the curriculum, and the automated curriculum generating methods often assume known goal states or prior knowledge on how to manipulate the environment~\cite{florensa2017reverse, wang2019paired} and bias the exploration to a small subset of the tasks~\cite{sukhbaatar2018intrinsic}. Through implicitly designing a form of curriculum to first achieve easily attainable goals and then progress towards more difficult goals, Hindsight Experience Replay (HER) is the first work that allows complicated manipulation behaviors to be learned from scratch with only binary rewards~\cite{andrychowicz2017hindsight}. However, when the actual task goal is very distinct from what random policies can achieve, HER's effect is limited. As mentioned in~\citep{andrychowicz2017hindsight}, HER is unable to allow manipulators to learn pick-and-place tasks without using demonstration states during training. 
% It augments reward signals through replaying experiences with different goals than the original one to be attained, which implicitly defines a form of curriculum where the agent first learns to reach goals that are easily attainable with random exploration, allowing it to progressively move towards more difficult goals. However, according to~\citet{andrychowicz2017hindsight}, HER is still unable to allow the manipulator to learn pick-and-place tasks without using demonstration states during training. The same conclusion is also drawn from the experiments in this paper comparing the performance of HER and our empowerment-based approach on a pick-and-place task.

Hierarchical reinforcement learning (HRL) approaches utilize temporal abstraction~\cite{bacon2017option} or information asymmetry~\cite{galashov2018information, goyal2018transfer} to introduce inductive biases for learning complicated tasks and transferable skills.
% divide the policy into different levels in order to better concentrate on different tasks during learning~\cite{daniel2016hierarchical, kulkarni2016hierarchical, haarnoja2018latent}.
% Temporal abstraction has been applied in the option-critic architecture~\cite{bacon2017option} to scale up learning and planning for temporally extended tasks, information asymmetry between different policy levels and probabilistic modeling of the objective function~\cite{galashov2018information, tirumala2019exploiting, goyal2018transfer} have been utilized to introduce inductive biases for learning complicated tasks and transferable skills, mutual information maximization~\cite{osa2018hierarchical, eysenbach2018diversity} has been applied to find a diverse set of action modes, 
Frameworks that combine multiple different tasks through a high level task selection policy~\cite{riedmiller2018learning, colas2019curious} have also shown effectiveness for learning sparse reward tasks. 
% However, many of the existing HRL approaches only enable the robot to accomplish one single task without the ability to transfer the skills to diverse task sets, and among the methods that automatically learn skills, it is still not clear whether a useful and diverse set of skills can be learned efficiently in complicated robotic manipulation tasks with sparse rewards.
Intrinsic exploration approaches, instead, augments the reward signals by adding task-agnostic rewards which encourage the agent to explore novel or uncertain states~\cite{kim2019curiosity}. Many approaches in the theme of intrinsic exploration have been proposed to alleviate the burden of reward engineering when training RL agents: visit counts and pseudo-counts~\cite{tang2017exploration} encourage the agent to explore states that are less visited; novelty-based approaches~\citep{pathak2017curiosity, pathak2019self} motivate the agent to conduct actions that lead to more uncertain results; reachability-based approaches~\citep{savinov2018episodic} gives rewards to the observations outside of the explored states that take more environment steps to reach; diversity-driven approaches~\citep{eysenbach2018diversity, sharma2020dynamics} learn skills using a maximum entropy policy to allow for the unsupervised emergence of diverse skills; and information gain~\citep{mohamed2015variational, houthooft2016vime, kim2019emi} encourages the agent to explore states that will improve its belief about the dynamics. However, count-based and uncertainty-based exploration methods often can't distinguish between task-irrelevant distractions and task-related novelties, and the high computational complexity largely restricts the application of existing information-theoretic methods in practical robotic manipulation tasks. The approach proposed in this paper falls under the category of information-theoretic intrinsic exploration, and we provide insight into reasonable approximations that can make the computation of information-theoretic quantities feasible when the state and action spaces are continuous and high-dimensional with complex mutual dependencies. Extensive experiment results demonstrate the effectiveness of these approximations as well as the superiority of the proposed approach over existing intrinsic exploration approaches in robotic manipulation scenarios.
% To the authors' best knowledge, there hasn't been successful demonstrations of intrinsic exploration in terms of facilitating the learning process of robotic manipulation tasks with sparse rewards. Therefore, in this paper, we are filling in this gap by providing extensive experiment results comparing the performance of three different intrinsic exploration approaches on object lifting tasks.

\section{Preliminaries} \label{sec:preliminaries}
   
\subsection{Mutual Information} \label{sec:MI}

\paragraph{Definition}
Mutual information (MI) is a fundamental quantity for measuring the mutual dependence between random variables. It quantifies the amount of information obtained about one random variable through observing the other. For a pair of continuous variables $X$ and $Y$, MI is defined as:

\begin{equation}
\label{equ:MI_def}
\begin{aligned}
 \mathcal{I}(X; Y) & = \iint p_{XY}(x, y) \log \frac{p_{XY}(x, y)}{p_{X}(x) p_{Y}(y)} \,dx\,dy \\
 & = \mathbb{E}_{XY} \Big[ \log \frac{p_{XY}}{p_{X} p_{Y}} \Big],
\end{aligned}
\end{equation}

\noindent where $p_{X}(x)$ and $p_{Y}(y)$ are the marginal probability density functions for $X$ and $Y$ respectively, and $p_{XY}(x, y)$ is the joint probability density function. MI is also often expressed in terms of Shannon entropy~\cite{paninski2003estimation} as well as Kullback-Leibler (KL) divergence:

\begin{equation}
\label{equ:Entropy}
\begin{aligned}
 \mathcal{I}(X; Y) & = \mathcal{H}(X) - \mathcal{H}(X|Y)  = \mathcal{H}(Y) - \mathcal{H}(Y|X) \\ 
 & = D_{KL}(p_{XY}||p_X p_Y), \\
\end{aligned}
\end{equation}

\noindent where $\mathcal{H}(X)$ and $\mathcal{H}(Y)$ are the marginal entropies, $\mathcal{H}(X|Y)$ and $\mathcal{H}(Y|X)$ are conditional entropies, $\mathcal{H}(X, Y)$ is the joint entropy, and $D_{KL}(p_{XY}||p_X p_Y)$ denotes the KL-divergence between the joint distribution and the product of the marginal distributions.

Conditional MI measures the mutual dependency between two random variables conditioned on another random variable. For continuous variables $X$, $Y$ and $Z$, conditioned MI is defined as:

\begin{equation}
\label{equ:CondMI}
 \begin{aligned}
 \mathcal{I}(X; Y| Z) & = \iiint \log \Big( \frac{p_{XY|Z}(x, y|z)}{p_{X|Z}(x|z)p_{Y|Z}(y|z)} \Big) \cdot \\
 & ~~~~~~~~~~~~~ p_{X,Y,Z}(x, y, z) \,dx \,dy \,dz \\
 & = \mathbb{E}_{XY|Z} \Big[ \log \frac{p_{XY|Z}}{p_{X|Z} p_{Y|Z}} \Big], 
\end{aligned}
\end{equation}

\noindent where $p_{X,Y,Z}(x, y, z)$ is the joint probability density function, and $p_{X,Y|Z}(x, y|z)$, $p_{X|Y,Z}(x|y, z)$, $p_{Y|X,Z}(y|x, z)$, $p_{X|Z}(x|z)$ and $p_{Y|Z}(y|z)$ are conditional probability density functions.

\paragraph{Computation}
In general, the computation of MI is intractable. Exact computation of MI is only tractable for discrete random variables and a limited family of problems with known probability distributions~\cite{belghazi2018mutual}. Traditional algorithms for MI maximization, e.g. the Blahut-Arimoto algorithm~\cite{cover2012elements}, don't scale well to realistic problems because they typically rely on enumeration. Therefore, researchers often maximize a lower bound of MI instead of computing its exact value. 

The variational lower bound derived from the non-negativity of KL-divergence, shown in Equation~\ref{equ:VariationalLowerBound}, is one of the most commonly used lower bounds for MI in the RL community:

% \begin{equation}
% \label{equ:VariationalLowerBound}
%  \begin{aligned}
%   \mathcal{I}(X; Y) & = \mathbb{E}_{XY} \Big[ \log \frac{p(x|y) \cdot q(x|y)}{p(x) \cdot q(x|y)} \Big] = \mathbb{E}_{XY} \Big[ \log \frac{q(x|y)}{p(x)} \Big] + \mathbb{E}_{XY} \Big[ \log \frac{p(x|y)}{q(x|y)} \Big] \\
%   & = \mathbb{E}_{XY} \Big[ \log \frac{q(x|y)}{p(x)} \Big] + \mathbb{E}_Y \Big[ D_{KL} (p(x|y) || q(x|y)) \Big] \geq \mathbb{E}_{XY} \Big[ \log \frac{q(x|y)}{p(x)} \Big], \\
%  \end{aligned}
% \end{equation}

\begin{equation}
\label{equ:VariationalLowerBound}
\begin{aligned}
  \mathcal{I}(X; Y) & = \mathbb{E}_{XY} \Big[ \log \frac{q(x|y)}{p(x)} \Big] + \mathbb{E}_Y \Big[ D_{KL} (p(x|y) || q(x|y)) \Big] \\
  & \geq \mathbb{E}_{XY} \Big[ \log \frac{q(x|y)}{p(x)} \Big], 
\end{aligned}
\end{equation}

\noindent where $q(x|y)$ is a variational approximation of $p(x|y)$, and the bound is tight when $q(x|y) = p(x|y)$. 
% For conditional MI $\mathcal{I}(X; Y| Z)$, the variational lower bound can be derived as:
% 
% \begin{equation}
% \label{equ:ConditionalVariationalLowerBound}
% % \small
%  \begin{aligned}
%   \mathcal{I}(X; Y|Z) & = \mathbb{E}_{XY|Z} \Big[ \log \frac{p(x|y, z) \cdot q(x|y, z)}{p(x|z) \cdot q(x|y, z)} \Big] = \mathbb{E}_{XY|Z} \Big[ \log \frac{q(x|y, z)}{p(x|z)} \Big] + \mathbb{E}_{XY|Z} \Big[ \log \frac{p(x|y, z)}{q(x|y, z)} \Big] \\
%   & = \mathbb{E}_{XY|Z} \Big[ \log \frac{q(x|y, z)}{p(x|z)} \Big] + \mathbb{E}_{Y|Z} \Big[ D_{KL} (p(x|y, z) || q(x|y, z)) \Big] \geq \mathbb{E}_{XY|Z} \Big[ \log \frac{q(x|y, z)}{p(x|z)} \Big]. \\
%  \end{aligned}
% \end{equation}

Other variational lower bounds of MI have also been derived based on a broader class of distance measures called $f$-divergence~\cite{liese2006divergences, nguyen2010estimating, nowozin2016f}. 
% The variational lower bound of $f$-divergences has been derived in \cite{} and \cite{}:
% \begin{equation}
% \label{equ:f_bound}
%  D_{f}(P(z)||Q(z)) \geq \sup_{T \in \mathcal{T}} (\mathbb{E}_{z \sim P}[T(z)] - \mathbb{E}_{z \sim Q}[f^* (T(z))]),
% \end{equation}
% \noindent where $\mathcal{T}$ is and arbitrary class of functions $T: \mathcal{Z} \rightarrow \mathbb{R}$, and $f^*$ is the convex conjugate of $f$. KL-divergence is a special case of $f$-divergence when the generator function $f(u) = u \log u$, and Jensen-Shannon (JS) divergence is a special case when $f(u)=-(u+1)\log((1+u)/2) + u\log u$~\cite{nowozin2016f}.
KL-divergence and Jensen-Shannon (JS) divergence are two special cases of $f$-divergence.
Based on the relationship between MI and KL-divergence shown in Equation~\ref{equ:Entropy}, a lower bound of MI is derived in~\cite{belghazi2018mutual}:

\begin{equation}
\label{equ:KLD}
 \mathcal{I}_{KL}(X; Y) \geq \sup_{T \in \mathcal{T}} \mathbb{E}_{p_{XY}} [T] - \mathbb{E}_{p_{X} p_{Y}}[e^{T - 1}],
\end{equation}

\noindent where $\mathcal{T}$ is an arbitrary class of functions $T: \mathcal{X} \times \mathcal{Y} \rightarrow \mathbb{R}$. 
% For conditional MI $\mathcal{I}(X; Y| Z)$, the lower bound can be written as:
% 
% \begin{equation}
% \label{equ:ConditionalKLD}
%  \mathcal{I}_{KL}(X; Y|Z) \geq \sup_{T \in \mathcal{T}} \mathbb{E}_{p_{XY|Z}} [T] - \mathbb{E}_{p_{X|Z} p_{Y|Z}}[e^{T - 1}].
% \end{equation}
The JS definition of MI is closely related to the MI we defined in Equation~\ref{equ:MI_def}, and its lower bound can be derived as~\cite{kim2019emi}:

% \begin{equation}
% \label{equ:JSD}
% \begin{aligned}
%  \mathcal{I}_{JS}(X; Y) & = D_{JS} (p_{XY}||p_X p_Y) \\
%  & \geq \sup_{T \in \mathcal{T}} \mathbb{E}_{p_{XY}} [\log 2 - \log (1 + e^{-T})] - \mathbb{E}_{p_X p_Y} [D_{JS}^* (\log 2 - \log (1 + e^{-T}))] \\
%  & = \sup_{T \in \mathcal{T}} \mathbb{E}_{p_{XY}} [-\textrm{sp} (-T)] - \mathbb{E}_{p_X p_Y} [\textrm{sp} (T)] + \log 4, \\
% \end{aligned}
% \end{equation}

\begin{equation}
\label{equ:JSD}
\begin{aligned}
 \mathcal{I}_{JS}(X; Y) & = D_{JS} (p_{XY}||p_X p_Y) \\
 & \geq  \sup_{T \in \mathcal{T}} \mathbb{E}_{p_{XY}} [-\textrm{sp} (-T)] - \mathbb{E}_{p_X p_Y} [\textrm{sp} (T)] + \log 4, 
\end{aligned}
\end{equation}

\noindent where $D_{JS}^* (u) = -\log (2-\exp(u))$ is the Fenchel conjugate of JS-divergence, and $\textrm{sp} (u) = \log (1+\exp (u))$ is the soft plus function. Detailed derivations of the above bounds and their counterparts for conditional MI estimation are provided in Appendix~\ref{sec:LowerBoundDerivation}. Note that Equation~\ref{equ:JSD} is not a lower bound for the MI we defined in Equation~\ref{equ:MI_def}, but since the two MIs are closely related, it is also often used to estimate the MI defined in Equation~\ref{equ:MI_def}. In this paper, we refer to the variational lower bound in Equation~\ref{equ:VariationalLowerBound} as VLB, the lower bound based on KL-divergence in Equation~\ref{equ:KLD} as KLD, and the lower bound for JS-divergence based mutual information in Equation~\ref{equ:JSD} as JSD. 
% The JSD lower bound for conditional MI can be written as:
% 
% \begin{equation}
% \label{equ:ConditionalJSD}
% \begin{aligned}
%  \mathcal{I}_{JS}(X; Y|Z) & = D_{JS} (p_{XY|Z}||p_{X|Z} p_{Y|Z}) \\
%  & \geq \sup_{T \in \mathcal{T}} \mathbb{E}_{p_{XY|Z}} [\log 2 - \log (1 + e^{-T})] - \mathbb{E}_{p_{X|Z} p_{Y|Z}} [D_{JS}^* (\log 2 - \log (1 + e^{-T}))] \\
%  & = \sup_{T \in \mathcal{T}} \mathbb{E}_{p_{XY|Z}} [-\textrm{sp} (-T)] - \mathbb{E}_{p_{X|Z} p_{Y|Z}} [\textrm{sp} (T)] + \log 4, \\
% \end{aligned}
% \end{equation}
% 
% \noindent where $\mathcal{T}$ is an arbitrary class of functions $T: \mathcal{X} \times \mathcal{Y} \times \mathcal{Z} \rightarrow \mathbb{R}$.

\subsection{Markov Decision Process (MDP)}

The problem studied in this paper is formulated as a Markov Decision Process (MDP) defined by states $\mathbf{s} \in \mathcal{S}$, actions $\mathbf{a} \in \mathcal{A}$, a transition model $T: \mathcal{S} \times \mathcal{A} \times \mathcal{S} \rightarrow \mathbb{R}$, and a reward function $r: \mathcal{S} \times \mathcal{A} \rightarrow \mathbb{R}$. $\mathcal{S}$ and $\mathcal{A}$ represent the state space and the action space respectively. The objective of the RL problem is to find a policy $\pi: \mathcal{S} \rightarrow \mathcal{A}$ that maximizes $J=\mathbb{E}_{\pi}[\sum_{\tau} r(\mathbf{s}_t, \mathbf{a}_t) | \mathbf{a}_t \sim \pi (\mathbf{s}_t), \mathbf{s}_0 \sim p_0(\mathbf{s})]$, where $\tau$ denotes the trajectory. In this paper, we refer to the reward from the environment as extrinsic reward $r^e$ and the artificial reward from the algorithm as intrinsic reward $r^i$, hence $r = r^e + r^i$. The sum $r$ is used during the learning process, whereas only $r^e$ is considered when evaluating the performance of a learning algorithm.
% The discount factor $\gamma \in [0, 1]$ describes the discount given to future rewards. 
% A policy $\pi: \mathcal{S} \rightarrow \mathcal{A}$ denotes a function on how to choose action at each state. 
% A typical objective of a RL problem is to maximize the sum of the discounted rewards $J=\mathbb{E}_{\pi}[\sum_{t=0}^{\infty} \gamma^t r(\mathbf{s}_t, \mathbf{a}_t) | \mathbf{a}_t \sim \pi (\mathbf{s}_t), \mathbf{s}_0 \sim p_0(\mathbf{s})]$.

\subsection{Empowerment} \label{sec:empowerment}

Empowerment is an information-theoretic quantity that measures the value of the information an agent obtains in the action-observation sequences it experiences during the reinforcement learning process~\cite{mohamed2015variational}. It is defined as the maximum mutual information between a sequence of $K$ actions $\mathbf{a}$ and the final state $\mathbf{s'}$, conditioned on a starting state $\mathbf{s}$:

\begin{equation}
\label{equ:empowerment}
\begin{aligned}
 \mathcal{E}(\mathbf{s}) & = \max_{\pi} \mathcal{I}^{\pi}(\mathbf{a}, \mathbf{s'} | \mathbf{s}) \\
 & = \max_{\pi} \mathbb{E}_{p(s'|a,s)\pi(a|s)}\bigg[ \log \bigg( \frac{p(\mathbf{a},\mathbf{s'}|\mathbf{s})}{\pi(\mathbf{a}|\mathbf{s})p(\mathbf{s'}|\mathbf{s})} \bigg)  \bigg],
\end{aligned}
\end{equation}

\noindent where $\mathbf{a} = \{a_1, \dots, a_K\}$ is a sequence of $K$ primitive actions leading to a final state $\mathbf{s'}$, $\pi(\mathbf{a}|\mathbf{s})$ is exploration policy over the $K$-step action sequences, $p(\mathbf{s'}|\mathbf{a}, \mathbf{s})$ is the $K$-step transition probability of the environment, $p(\mathbf{a},\mathbf{s'}|\mathbf{s})$ is the joint distribution of actions sequences and the final state conditioned on the initial state $\mathbf{s}$, and $p(\mathbf{s'}|\mathbf{s})$ is the marginalized probability over the action sequence. 
% Through measuring the mutual dependence between actions and states, empowerment indicates how confident the agent is about the effect of its actions in the environment. In contrast to novelty-driven intrinsic motivations which encourage the agent to explore actions with unknown effects, empowerment emphasizes the agent's ``controllability'' over the environment and favors actions with predictable consequences. We hypothesize that empowerment is a more suitable intrinsic motivation for robotic manipulation tasks where the desired interactions with environment objects are typically predictable and principled. Imaging a robot is interacting with a box on the table. Intuitively, the behaviors of knocking the box onto the floor should generate higher novelty since it helps explore more states that haven't been visited, and the behaviors of pushing the box or lifting the box up should generate higher empowerment because the effects of these actions are more predictable. Based on this intuition, we propose an empowerment-based intrinsic motivation in this paper and provide extensive experiment results in simulation environments to demonstrate its performance.

\subsection{Intrinsic Curiosity Module}

Intrinsic Curiosity Module (ICM)~\cite{pathak2017curiosity} is one of the state-of-the-art novelty-driven intrinsic exploration approaches that aims at learning new skills by performing actions whose consequences are hard to predict. It trains an inverse model $g$ to learn a feature encoding $\phi$ that captures the parts of the state space related to the consequences of the agent's actions, so that the agent will focus on the relevant part of the environment and not get distracted by other details in the camera observations. It also learns the forward model $f$ and uses the prediction error of the forward model as the intrinsic reward in order to facilitate the agent to explore the part of the state space where it can't predict the consequences of its own actions very well. 

% \begin{equation}
%   \mbox{Inverse Model:~} \hat{a}_t = g(\phi(s_t),\phi(s_{t+1});\theta_I); 
%   ~~~\mbox{Forward Model:~} \hat{\phi}(s_{t+1}) = f(\phi(s_t),a_t;\theta_F). 
% \end{equation}

\begin{equation}
\begin{aligned}
  & \mbox{Inverse Model:~} \hat{a}_t = g(\phi(s_t),\phi(s_{t+1})); \\
  & \mbox{Forward Model:~} \hat{\phi}(s_{t+1}) = f(\phi(s_t),a_t). 
\end{aligned}
\end{equation}

% The following can be moved to approach section
% In this paper, it is assumed that we can obtain robot and object poses and camera observations are not utilized, thus the feature encoding and the inverse model is no longer relevant. In addition, in manipulation tasks, since the transition model for robot states is trivial and the forward prediction error for robot states are not meaningful in terms of accomplishing object manipulation tasks, here we divide the state space $\mathbf{s}$ into intrinsic states $\mathbf{s}^{in}$ (all robot-related states) and extrinsic states $\mathbf{s}^{ex}$ (states that are only related to the object). In order to avoid having the robot gaining intrinsic reward simply by waving its arm around without touching the object, here we define the intrinsic reward as the forward prediction error of extrinsic states only:
% 
% \begin{equation}
% \begin{aligned}
%  & \hat{\mathbf{s}}_{t+1}^{ex} = f \big( \mathbf{s}_t^{ex}, \mathbf{a}_t \big), \\
%  & r_t^i = \frac{\eta}{2} || \hat{\mathbf{s}}_{t+1}^{ex} - \mathbf{s}_{t+1}^{ex} ||^2_2, \\
% \end{aligned}
% \end{equation}
% 
% \noindent where $\eta$ is the intrinsic reward coefficient.

\begin{figure*}
\centering
 \includegraphics[width=0.9\linewidth]{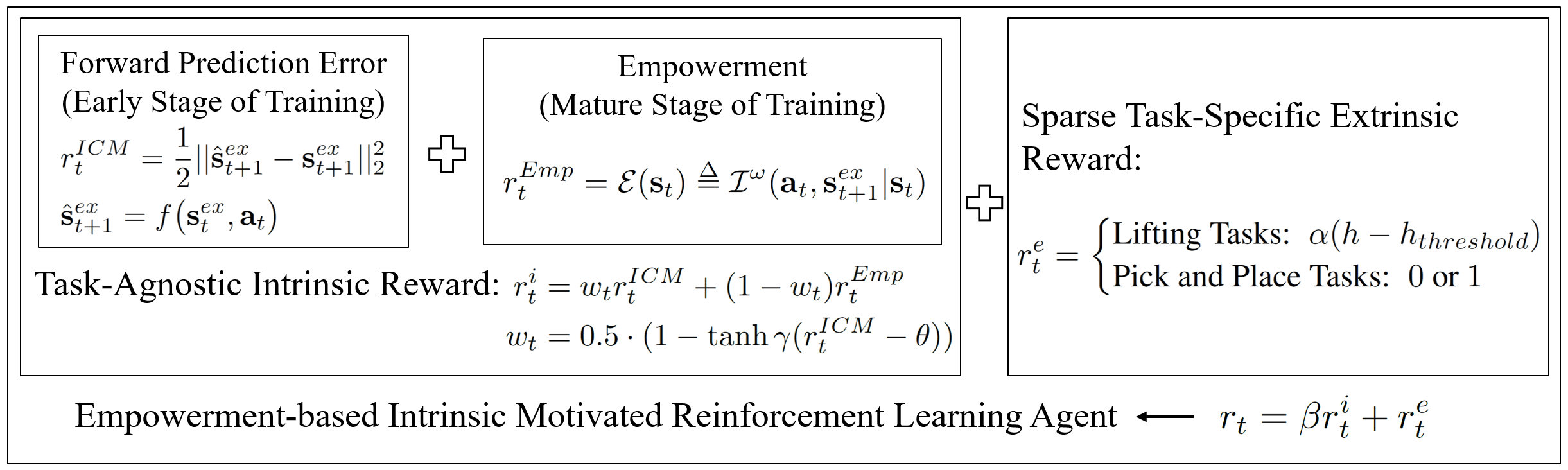}
 \caption{Overview of the empowerment-based intrinsic motivation approach}
\label{fig:diagram}
\end{figure*}

\section{Approach: Empowerment-based Intrinsic Motivation}

We hypothesize that empowerment would be a good candidate for augmenting the sparse extrinsic rewards in manipulation tasks because it indicates the amount of information contained in the action sequence $\mathbf{a}$ about the future state $\mathbf{s'}$. Through maximizing empowerment, we are effectively encouraging the agent to influence the environment in a predictable way, which is the desired behavior in most manipulation tasks. However, as a form of conditional MI for continuous variables, the computation of empowerment is especially challenging. This is because for conditional MI $\mathcal{I}(X; Y| Z)$ with continuous $Z$, estimating $\mathcal{I}(X; Y| Z)$ for all $Z$ is approximately equivalent to estimating an infinite number of unconditional MIs. In this section, we discuss the approaches we take to make empowerment a feasible form of intrinsic motivation in practical robotic manipulation tasks.

\subsection{Approximations to Simplify Empowerment Calculation}

\citet{mohamed2015variational} suggest that the empowerment at each state in the state space can be calculated using an exploration policy $\pi(\mathbf{a}|\mathbf{s})$ that generates an open-loop sequence of $K$ actions into the future (Equation~\ref{equ:empowerment}), so that a closed-loop policy can be obtained by a planning algorithm using the calculated empowerment values. Although \citeauthor{mohamed2015variational} demonstrated the effectiveness of this approach in grid world environments, it is infeasible to precompute the empowerment values for all states in a high-dimensional, continuous state space. Therefore, we make a few approximations in order to make empowerment-based intrinsic motivation a practical approach. First, we use only one action step instead of an action sequence to estimate empowerment. Second, instead of constructing a separate exploration policy $\pi$ to first compute empowerment and then plan a closed-loop policy according to empowerment, we directly optimize the behavior policy $\omega$ using empowerment as an intrinsic reward in an RL algorithm. These two approximations mean that the agent will only be looking at the one-step reachable neighborhood of its current state to find the policy that leads to high mutual information. Despite sacrificing global optimality, this approach prioritizes the policy that controls the environment in a principled way so that more extrinsic task rewards can be obtained compared to using random exploration, which help resolve the fundamental issue in sparse reward tasks.

In addition to the above two approximations, it is also important to note that in robotic manipulation tasks, we are typically not interested in the mutual dependence between robot actions and robot states, and we wish to avoid the robot trivially maximizing empowerment through motion of its own body. Therefore, we assume that the state space can be divided into intrinsic states $\mathbf{s}^{in}$ (robot states) and extrinsic states $\mathbf{s}^{ex}$ (environment states), and only extrinsic states are used as $\mathbf{s'}$ when calculating empowerment. Namely, the empowerment used in this paper is defined as:

\begin{equation}
 \label{equ:approx_empowerment}
 \mathcal{E}(\mathbf{s}_t) \approx \mathcal{I}^{\omega}(\mathbf{a}_t, \mathbf{s}^{ex}_{t+1} | \mathbf{s}_t) = \mathcal{H}^{\omega}(\mathbf{a}_t | \mathbf{s}_t) - \mathcal{H}^{\omega}(\mathbf{a}_t | \mathbf{s}^{ex}_{t+1}, \mathbf{s}_t),
\end{equation}

\noindent where $\omega$ is the behavior policy, and the relationship to Shannon Entropy is derived from Equation~\ref{equ:Entropy}.

\subsection{Maximizing Empowerment using Mutual Information Lower Bounds}

% As a form of conditional MI for continuous random variables, the computation of empowerment is especially challenging. This is because for conditional MI $\mathcal{I}(X; Y| Z)$ with continuous $Z$, estimating $\mathcal{I}(X; Y| Z)$ for all $Z$ is approximately equivalent to estimating an infinite number of unconditional MIs. 
% Closed form solutions for conditional MI are only available for a very small number of distributions. 
Neural function approximators have become powerful tools for numerically estimating conditional MIs for continuous random variables~\cite{mohamed2015variational, belghazi2018mutual, kim2019emi}. However, in most RL scenarios, since exact distributions are typically unavailable and numerical estimation through sampling is required, computation of high-dimensional conditional MI remain challenging.
As mentioned in Section~\ref{sec:MI}, a common practice is to maximize a lower bound of MI instead of its exact value. We test the performance of the three MI lower bounds introduced in Section~\ref{sec:MI} on distributions with known conditional MI and provide detailed experiment results in Appendix~\ref{sec:LowerBoundComparison}. We conclude that, in terms of estimating the conditional MI of the continuous random variables we tested on, VLB performs the best in all cases and KLD performs the worst in most cases. However, the same conclusion may not be drawn for high-dimensional distributions with complex mutual dependencies. In the manipulation tasks in this paper, we noticed that JSD is the best performer on Fetch and VLB is the best performer on PR2, hence we will report the results with the corresponding best performer in each environment.
% From the comparison between the RMSE and the absolute values of theoretical average MI we also conclude that it is possible to get a relatively accurate approximation of the conditional MI through numerical estimation when the mutual dependency between random variables are simple.

% \subsection{Reinforcement Learning with Empowerment-based Intrinsic Motivation}
\subsection{Combination with ICM to Facilitate Empowerment Computation}
% Even with the best performing lower bounds, in practice it is still very challenging to train neural networks to make accurate predictions on empowerment when the distributions are high-dimensional and have complicated dependencies. Therefore, we make a few approximations in order to make empowerment-based intrinsic motivation a practical approach. First, we use only one action step instead of an action sequence to estimate empowerment. Second, instead of constructing a separate exploration policy to first compute empowerment and then plan a closed-loop policy according to empowerment, we directly optimize the behavior policy by using empowerment as an intrinsic reward in a RL algorithm. 
% In addition, in robotic manipulation tasks, we are typically not interested in the mutual dependence between robot action and robot states, and we wish to avoid the robot trivially maximizing empowerment through motion of its own body. 

   \begin{figure*}[h!]
      \centering
      \begin{subfigure}[t]{0.25\textwidth}
      \centering
      \includegraphics[width=0.65\linewidth]{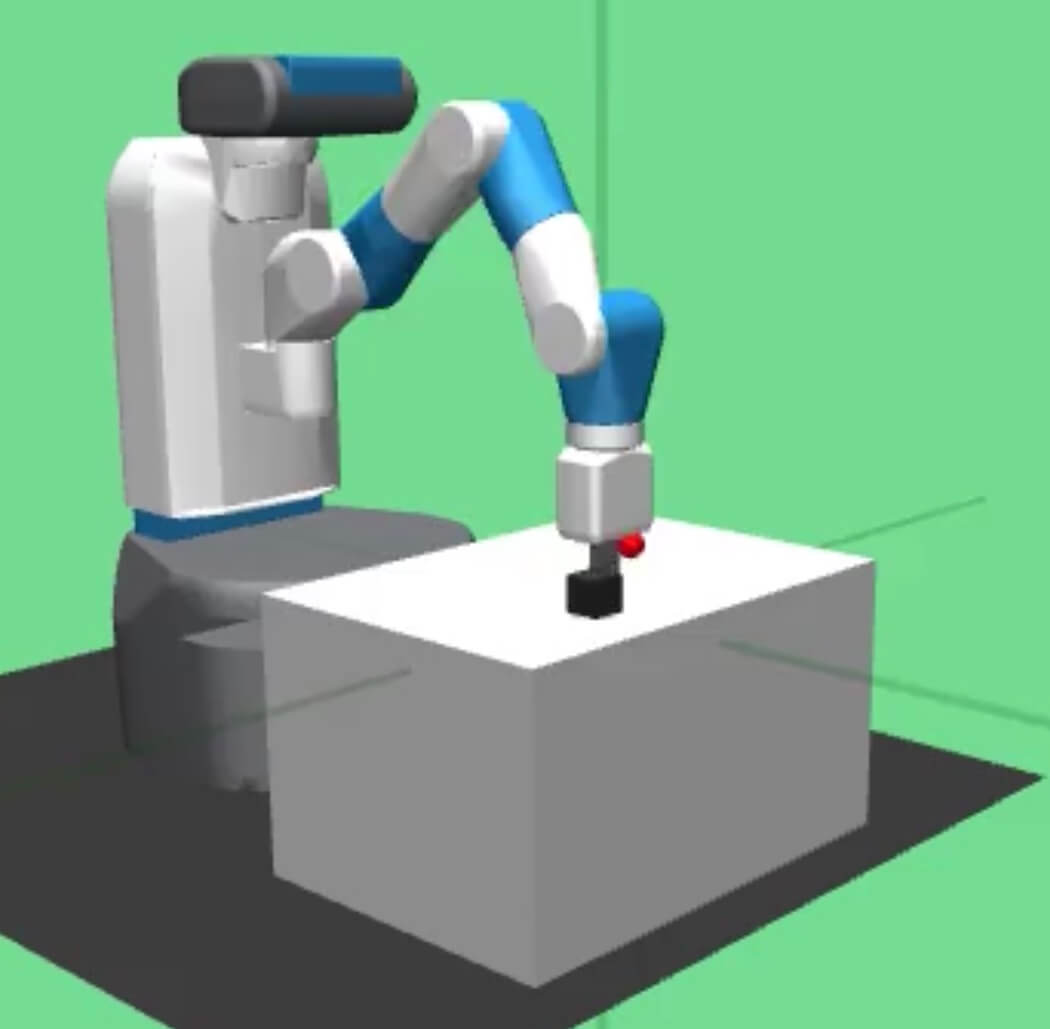}
      \caption{Fetch with a box}
      \end{subfigure}%
      \begin{subfigure}[t]{0.25\textwidth}
      \centering
      \includegraphics[width=0.65\linewidth]{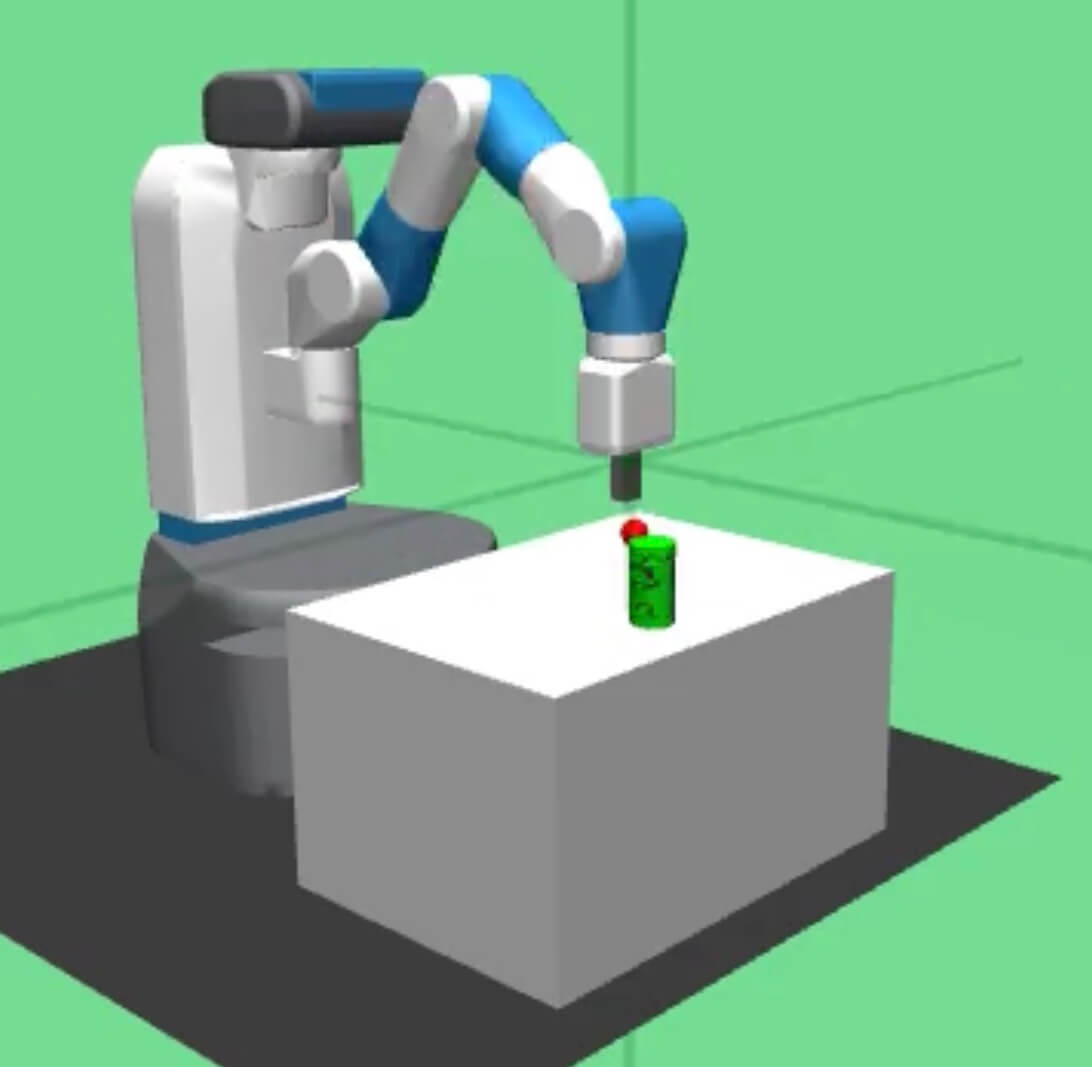}
      \caption{Fetch with a cylinder}
      \end{subfigure}%
      \begin{subfigure}[t]{0.25\textwidth}
      \centering
      \includegraphics[width=0.65\linewidth]{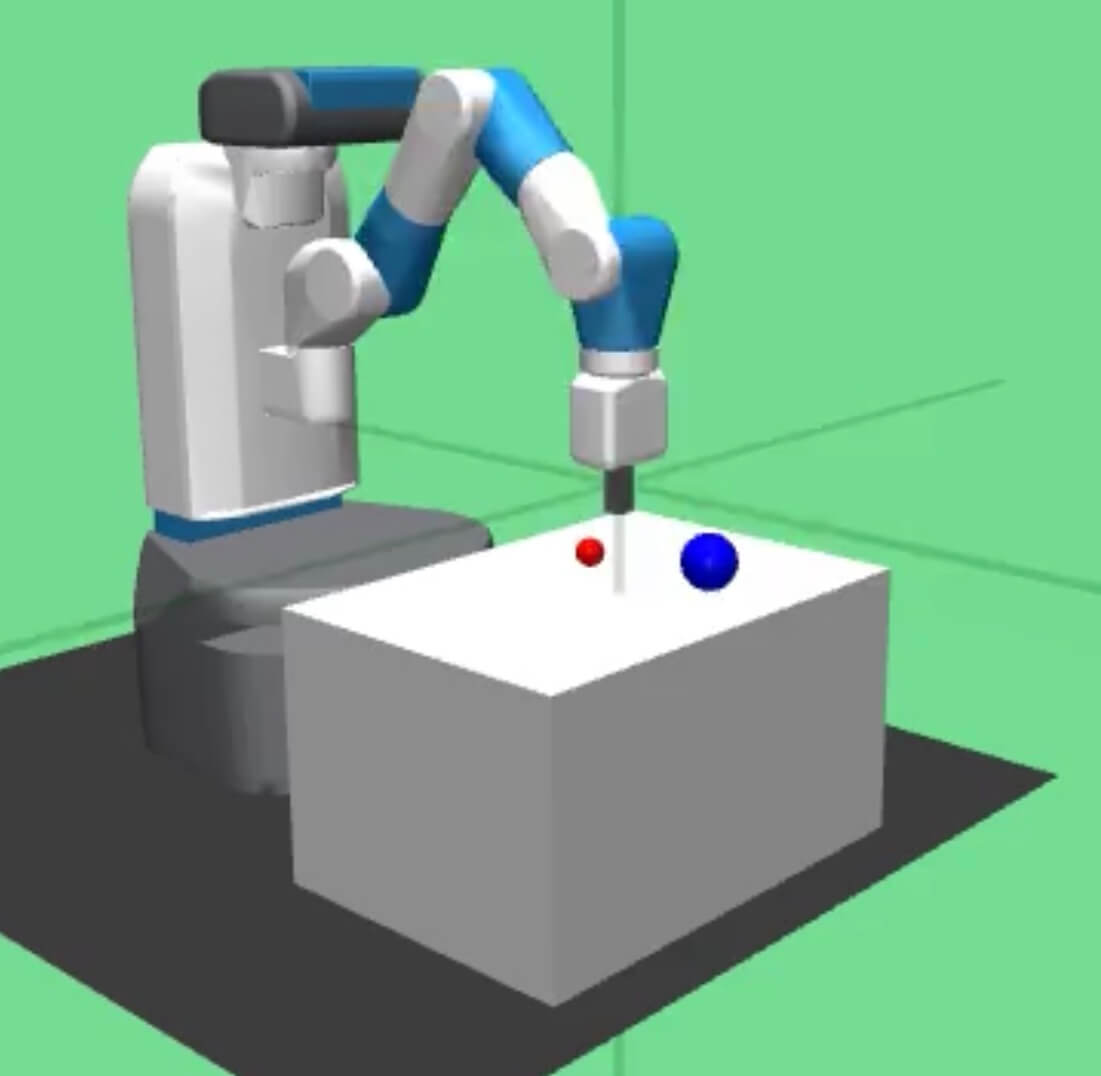}
      \caption{Fetch with a sphere}
      \end{subfigure}%
      \begin{subfigure}[t]{0.25\textwidth}
      \centering
      \includegraphics[width=0.65\linewidth]{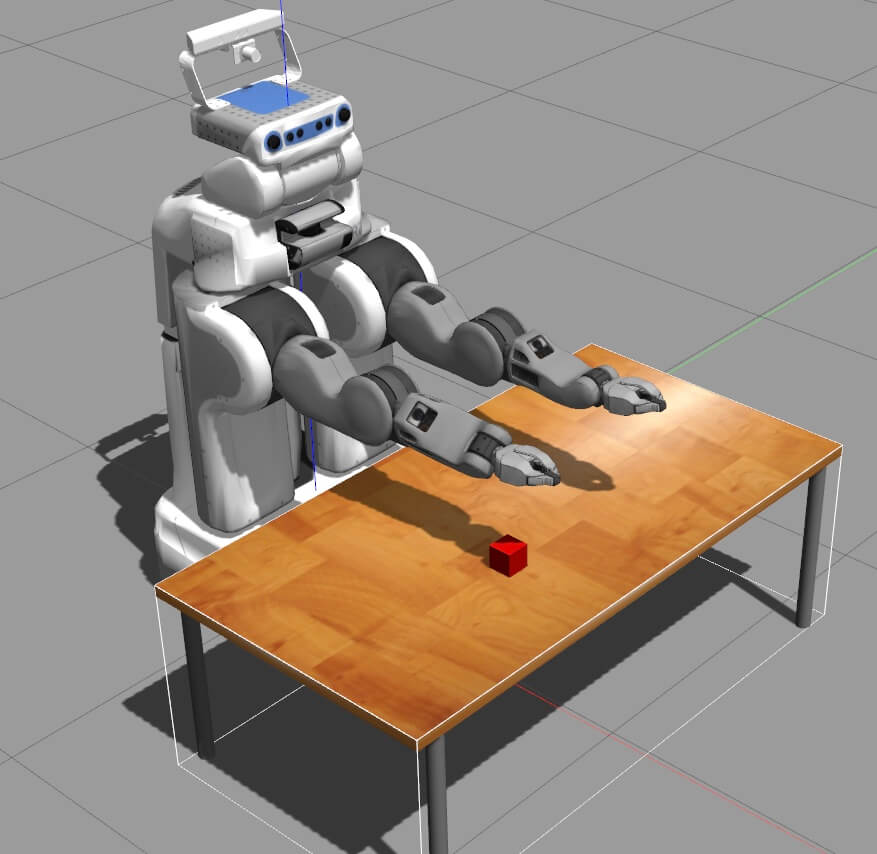}
      \caption{PR2 with a box}
      \end{subfigure}%
      \caption{Simulation environments}
      \label{fig:envs}
   \end{figure*}

   \begin{figure*}[h!]
      \centering
      \begin{subfigure}[t]{0.33\textwidth}
      \includegraphics[width=\linewidth]{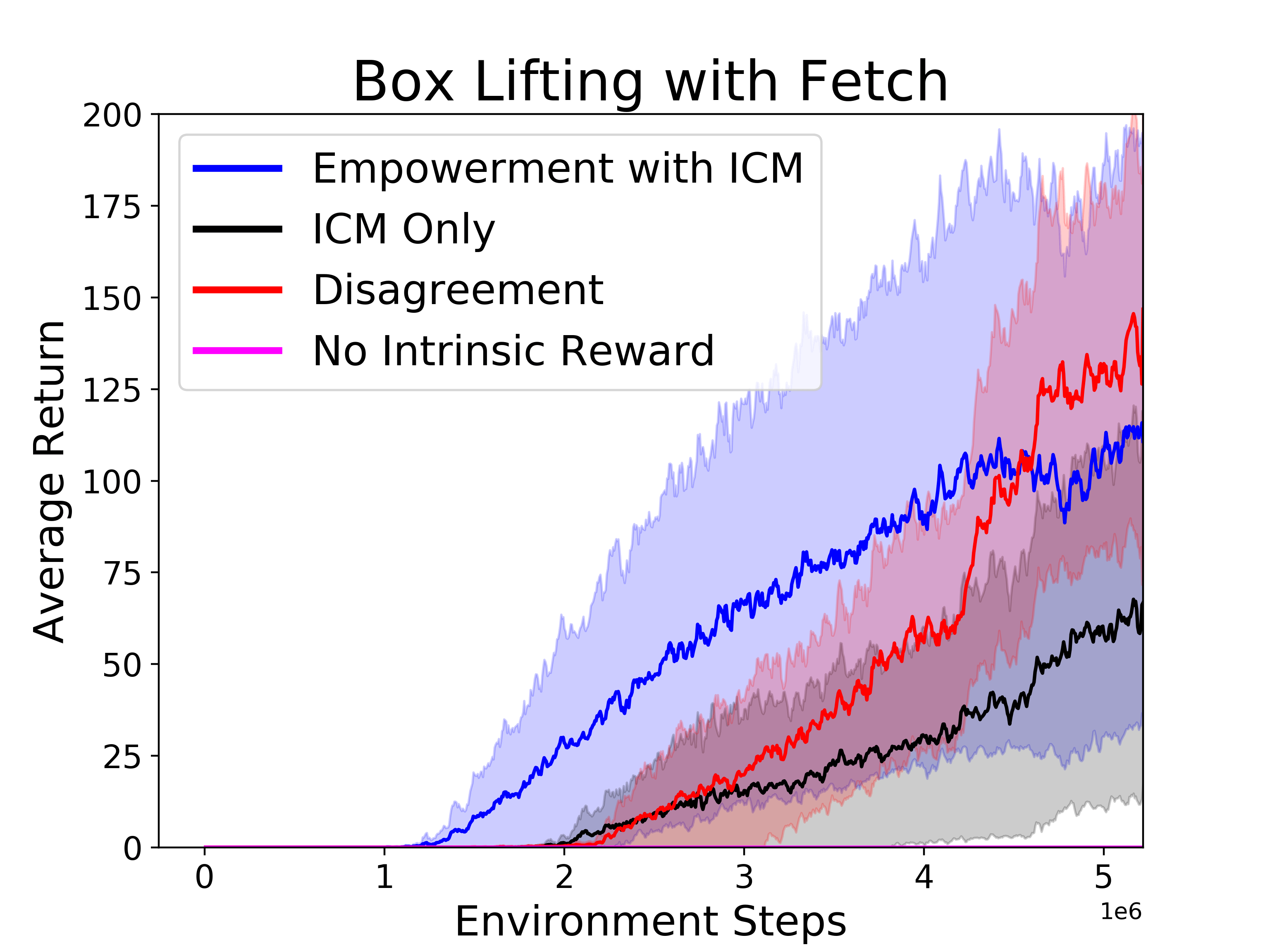}
      \caption{ }
      \end{subfigure}%
      \begin{subfigure}[t]{0.33\textwidth}
      \includegraphics[width=\linewidth]{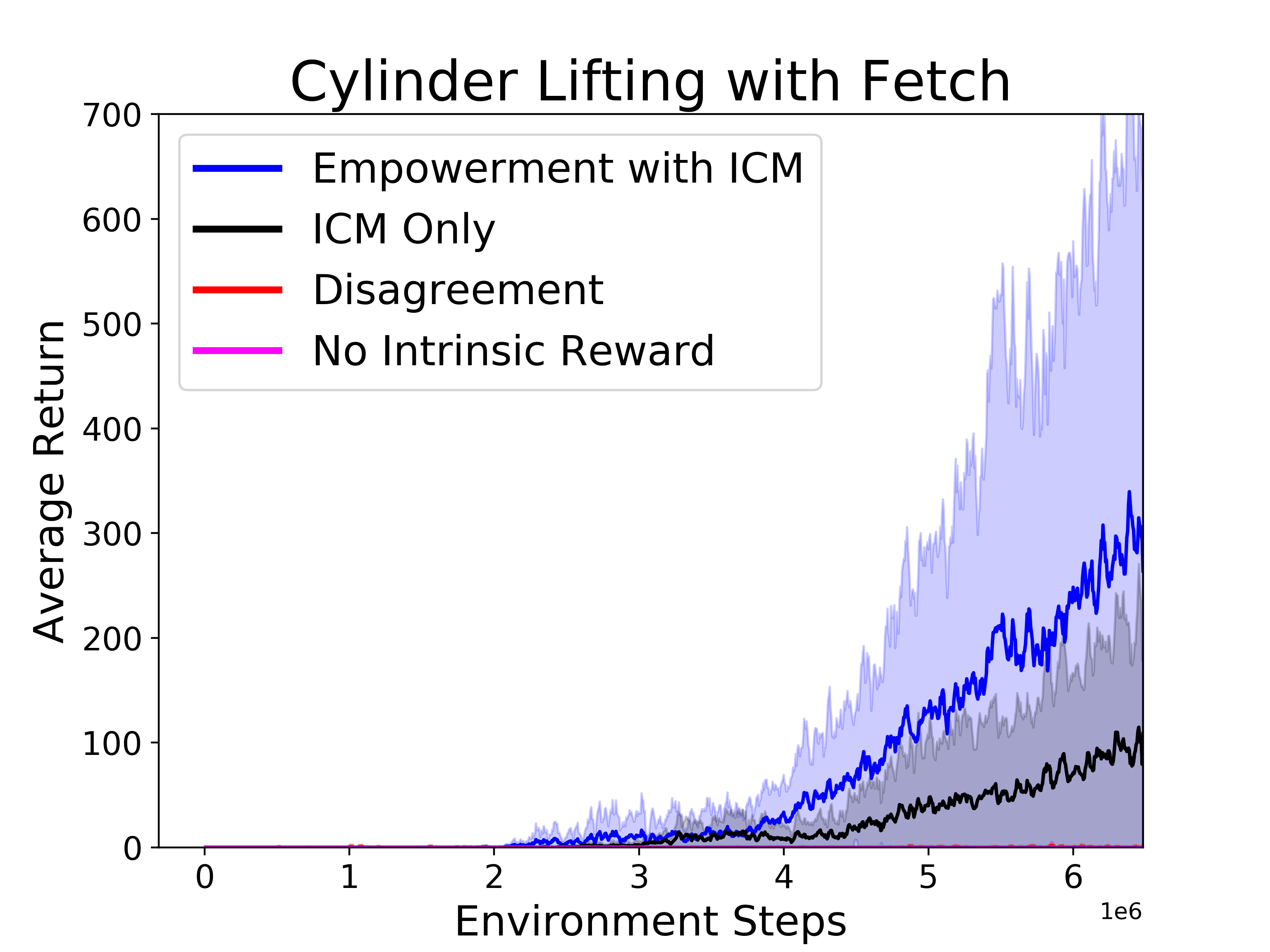}
      \caption{ }
      \end{subfigure}%
      \begin{subfigure}[t]{0.33\textwidth}
      \includegraphics[width=\linewidth]{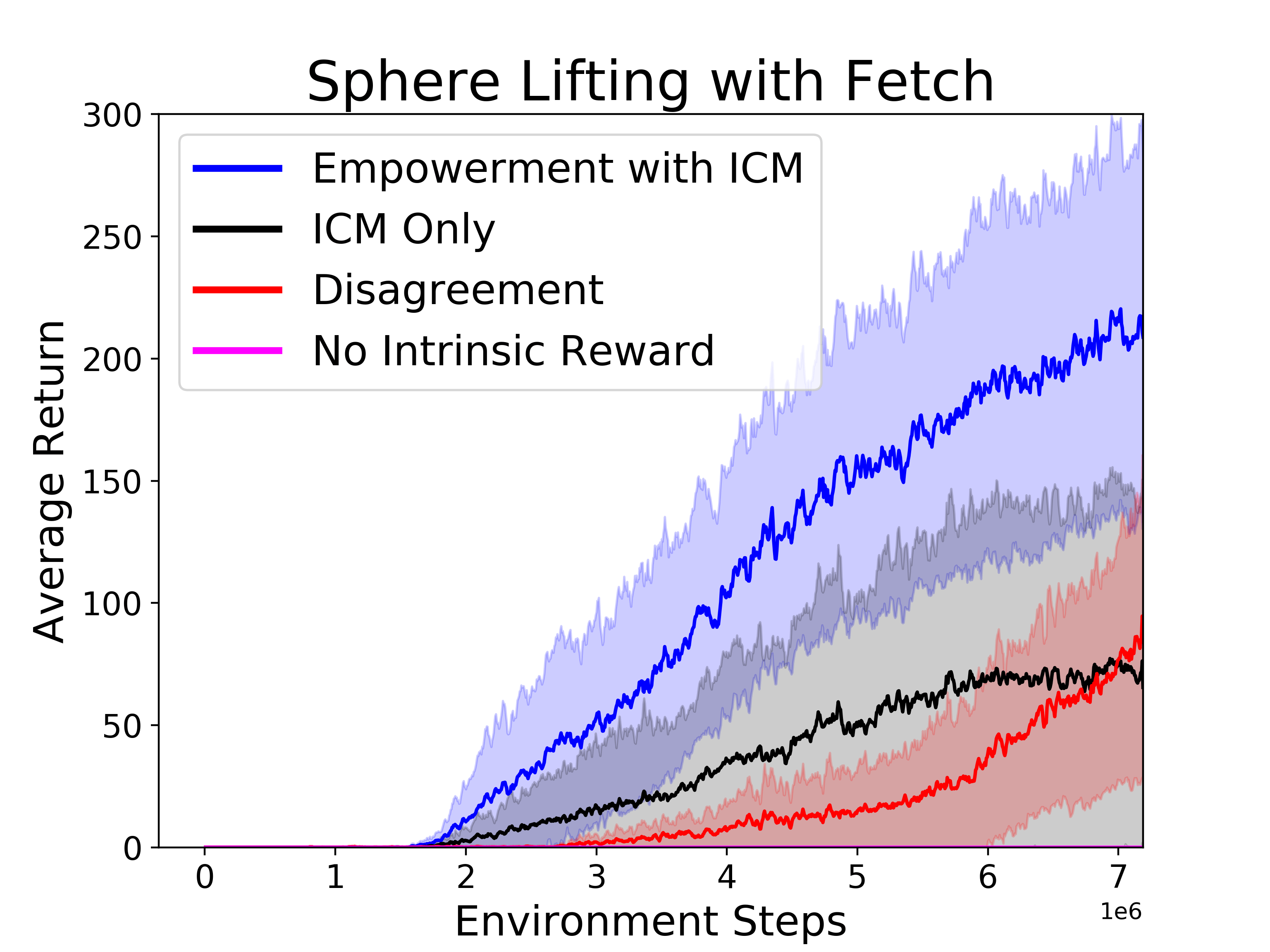}
      \caption{ }
      \end{subfigure}
      \begin{subfigure}[t]{0.33\textwidth}
      \includegraphics[width=\linewidth]{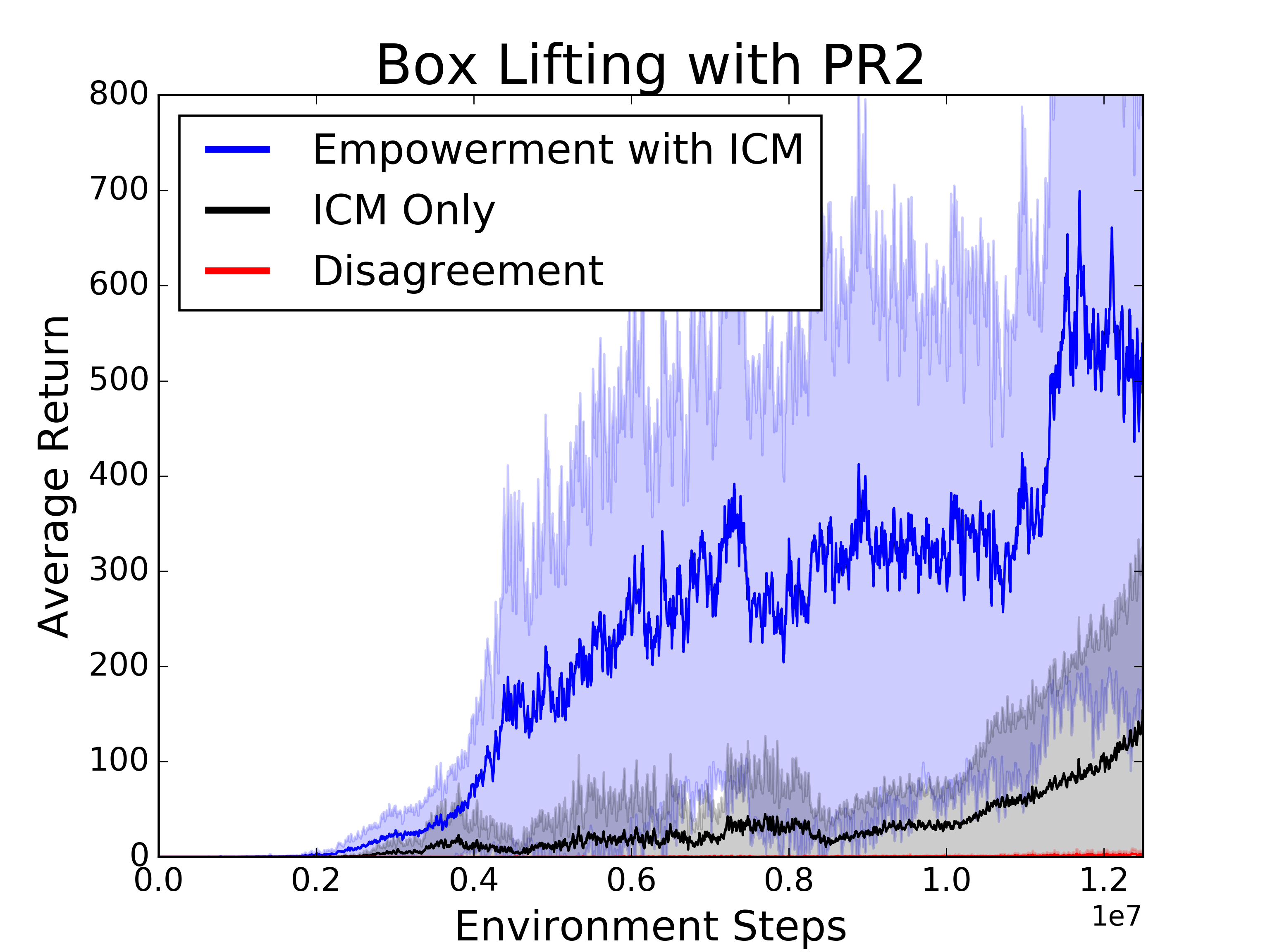}
      \caption{ }
      \end{subfigure}%
      \begin{subfigure}[t]{0.33\textwidth}
      \includegraphics[width=\linewidth]{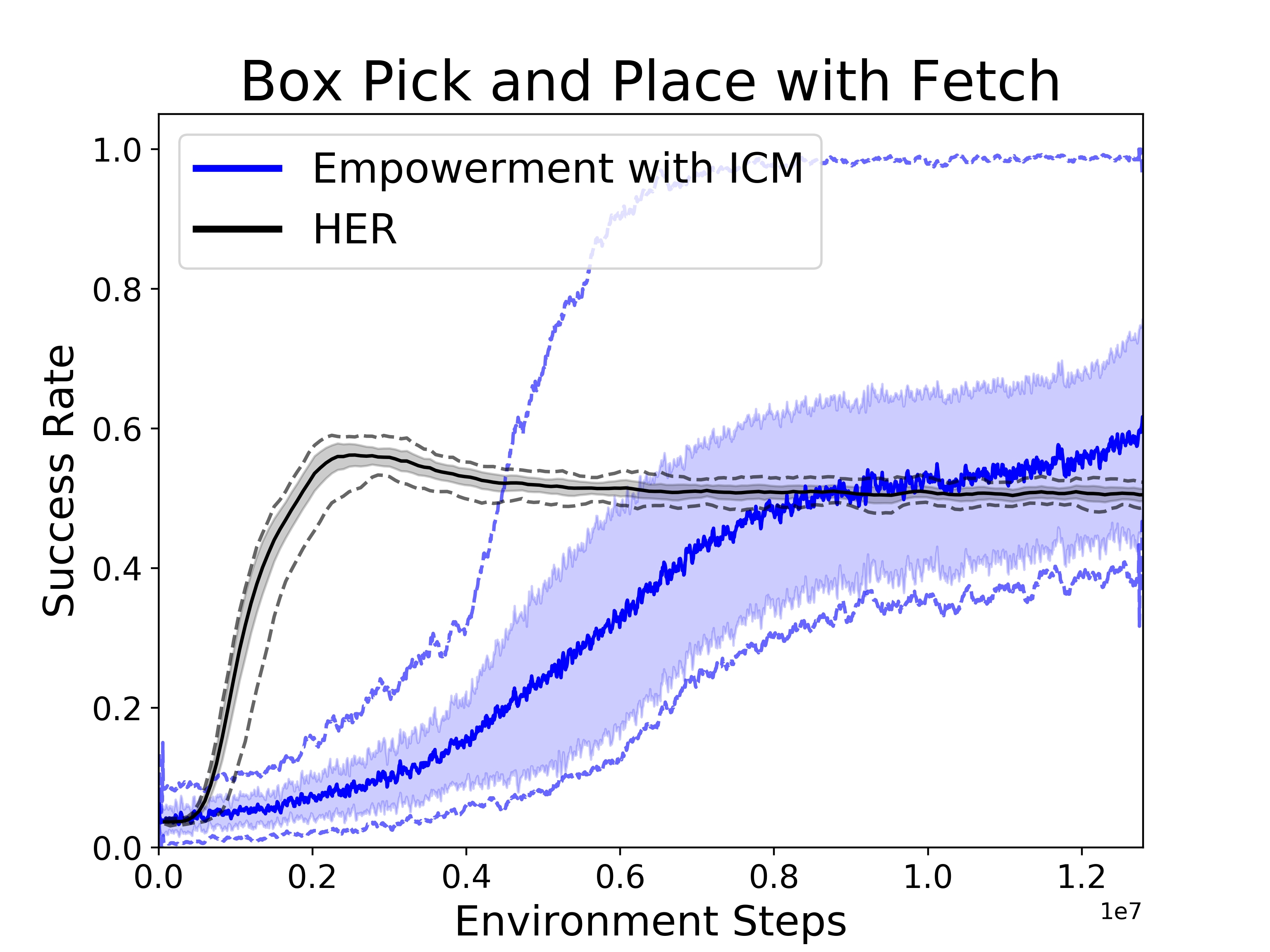}
      \caption{ }
      \end{subfigure}%
      \caption{Experiment results. (a)-(d) compare the performance of the proposed empowerment-based approach (referred to as empowerment with ICM since ICM is used to help training the empowerment prediction networks) with ICM and Disagreement in object lifting tasks, and (e) compares the proposed empowerment-based approach with HER in pick-and-place tasks. The solid lines represent the mean, the shadow areas represent the 95\% confidence intervals, and the dashed lines in (e) represent the maximum and minimum values.}
      \label{fig:return_comparison}
   \end{figure*}

Another challenging issue with empowerment-based RL is that well-balanced data are not easy to obtain at the beginning of training. If we initialize the RL agent with a random policy, it will highly likely explore much more of the empty space than regions with object interactions because the interaction-free part of the state space is often much larger. However, since $\mathbf{a}_t$ and $\mathbf{s}^{ex}_{t+1}$ are independent without interactions, the training data fed into the empowerment estimation network will be strongly biased towards the zero empowerment regions, which makes it very difficult to train accurate estimation models. Therefore, it is crucial that enough training data in the interacting part of the state space can be obtained at the beginning of training in order to get accurate estimations of empowerment. We achieve this through combining empowerment with the forward model of ICM using adaptive coefficients, which initially place more weight on ICM to ensure enough well-balanced data are fed to the empowerment estimation networks, and then switches more weight to empowerment to encourage the robot to learn controllable behaviors. Figure~\ref{fig:diagram} summarizes the proposed empowerment-based intrinsic motivation approach, and Appendix~\ref{sec:details} elaborates on the algorithm implementation details.

% \begin{equation}
%  \begin{aligned}
% %   & r_t^{ICM} = \frac{\eta}{2} || \hat{\mathbf{s}}_{t+1}^{ex} - \mathbf{s}_{t+1}^{ex} ||^2_2 \\
%   & r_t^{ICM} = \frac{1}{2} || \hat{\mathbf{s}}_{t+1}^{ex} - \mathbf{s}_{t+1}^{ex} ||^2_2 \\
%   & \hat{\mathbf{s}}_{t+1}^{ex} = f \big( \mathbf{s}_t^{ex}, \mathbf{a}_t \big) \\
%  \end{aligned}
% \end{equation}
% 
% \begin{equation}
% %  \begin{aligned}
%   r_t^{Emp} = \mathcal{E}(\mathbf{s}_t) \delequal \mathcal{I}^{\omega}(\mathbf{a}_t, \mathbf{s}^{ex}_{t+1} | \mathbf{s}_t) 
% %   & = \mathcal{H}^{\omega}(\mathbf{a}_t | \mathbf{s}_t) - \mathcal{H}^{\omega}(\mathbf{a}_t | \mathbf{s}^{ex}_{t+1}, \mathbf{s}_t) \\
% %  \end{aligned}
% \end{equation}
% 
% \begin{equation}
%  \begin{aligned}
%   r_t^i & = w_t r_t^{ICM} + (1-w_t) r_t^{Emp} \\
%   w_t & = 0.5 \cdot (1 -\tanh \gamma (r_t^{ICM} - \theta)) \\
%  \end{aligned}
% \end{equation}
% 
% % \begin{equation}
% %  \begin{aligned}
% %   r_t^e = \begin{cases} \mbox{Lifting Tasks:~ } \alpha (h - h_{threshold}) \\
% %    \mbox{Pick and Place Tasks:~ } 1 ~\mbox{if object reached target region,}~ 0 ~\mbox{otherwise} \end{cases}
% %  \end{aligned}
% % \end{equation}
% \begin{equation}
%  \begin{aligned}
%   r_t^e = \begin{cases} \mbox{Lifting Tasks:~ } \alpha (h - h_{threshold}) \\
%    \mbox{Pick and Place Tasks:~ } 0 ~\mbox{or}~ 1 \end{cases}
%  \end{aligned}
% \end{equation}
% 
% \begin{equation}
%  r_t = \beta r_t^i + r_t^e
% \end{equation}

%===============================================================================

\section{Empirical Evaluation}
\label{sec:result}

\subsection{Environment Setup}

In order to compare the performance of the empowerment-based intrinsic motivation with other state-of-the-art intrinsic motivations, we created four object-lifting tasks with different object shapes in OpenAI Gym~\citep{1606.01540} and Gazebo, as shown in Figure~\ref{fig:envs}. The Gym environment uses a Fetch robot with a 25D state space (including the poses and velocities of the end-effector, the gripper and the object) and a 4D action space (including the actions of the end-effector and gripper), and the Gazebo environment uses a PR2 robot with a 38D state space (including the poses and velocities of all joints and the object) and an 8D action space (including the actions of manipulator joints). We also use the FetchPickAndPlace-V1 task provided in Gym in order to compare with HER because HER requires a goal-conditioned environment. In the four object-lifting tasks, the goal is to lift up the object, and the extrinsic reward is only given when the object's height is above a threshold. In the pick-and-place task, the reward is given when the distance of the object to the goal pose is within a threshold. Our approach can be easily integrated with any standard RL algorithm, but in this paper, we use Proximal Policy Optimization (PPO)~\citep{schulman2017proximal} as the RL agent for all experiments to demonstrate its performance. Experiments on the Fetch robot use 60 parallel environments for training, and PR2 experiments use 40 due to its higher CPU requirement. Implementation details including hyperparameters and task rewards are provided in Appendix~\ref{sec:details}.

% The experiments in this paper are executed on the Fetch robot in the OpenAI Gym environment~\citep{1606.01540}. We compare the performance of our empowerment-based approach with two other state-of-the-art intrinsic exploration method on Fetch robot with three different objects: a box, a cylinder and a sphere, as shown in Figure~\ref{fig:envs}. The task goal is to lift up the object, and the extrinsic reward is only given when the object's height is above a threshold and the gripper is in contact with the object. The magnitude of the extrinsic reward is in proportion to the object's height above the threshold. In addition, we also compared our approach with HER in the FetchPickAndPlace-V1 environment provided in Gym because HER requires a goal-conditioned environment.

\subsection{Experiment Results}

In this section, we provide experiment results that compare the proposed empowerment-based intrinsic motivation approach with other state-of-the-art algorithms, including ICM~\citep{pathak2017curiosity}, exploration via disagreement~\citep{pathak2019self} (referred to as Disagreement in this paper) and HER~\citep{andrychowicz2017hindsight}. We use our implementation of ICM and Disagreement, and use the OpenAI Baselines implementation~\citep{baselines} for HER. In both ICM and Disagreement, we also make the same assumption as in the empowerment implementation that the state space can be divided into intrinsic states and extrinsic states, and only the prediction error or variance of the extrinsic states contribute to the intrinsic rewards. We run HER with 2 MPI processes with 30 parallel environments each to make sure it is equivalent to the 60 parallel environments in other experiments. Other parameters for HER are set to default. All the results in the Fetch environment are averaged over 10 different random seeds, and the results in the PR2 environment are averaged over 8 random seeds.

Figure~\ref{fig:return_comparison}(a)-(c) compare the performance of our approach with ICM, Disagreement, and PPO without any intrinsic reward in the object-lifting tasks with a Fetch robot, and Figure~\ref{fig:return_comparison}(d) compares our approach with ICM and Disagreement in box-lifting tasks with a PR2 robot. In the Fetch environment, the cylinder lifting task is much more difficult compared to box lifting and sphere lifting, thus we use a larger scale $\alpha$ for extrinsic lifting reward. Similarly, we also use a larger $\alpha$ for the box-lifting task with the PR2 robot since this environment is much higher-dimensional and hence more difficult for an RL agent. From Figure~\ref{fig:return_comparison}(a)-(c) we can see that the reward curve for PPO without any intrinsic reward remains almost zero, which proves that sparse reward tasks are very challenging for vanilla RL algorithms. In all four environments, our empowerment-based approach is able to help the robot achieve higher lifting rewards faster than other approaches we compared with. The Disagreement approach is able to perform better in the box lifting task with the Fetch robot after training for a long time, but it performs much worse than the other two intrinsic motivations in the cylinder and sphere lifting tasks. Another finding from Figure~\ref{fig:return_comparison}(a)-(c) is that the advantage of the empowerment-based intrinsic motivation is much more obvious in the cylinder and sphere lifting tasks compared to the box lifting tasks. We hypothesize that this is because the ability of ``controlling'' the object is much more important when there are round surfaces, since these objects are more difficult to pick up and also more likely to randomly roll around when novelty is the only intrinsic motivation. In fact, in the cylinder lifting task, our empowerment-based intrinsic motivation is the only approach that allows the agent to learn to directly pick up the cylinder from the top without knocking it down first, whereas agents trained with ICM will knock down the cylinder and then pick up radially. In Figure~\ref{fig:return_comparison}(d), although the confidence intervals are wider due to the smaller number of runs, we can still get the similar conclusion that our approach shows the best performance.

Figure~\ref{fig:return_comparison}(e) compares the empowerment-based intrinsic motivation with HER in the Fetch pick-and-place environment. We can see that although the average success rate of HER goes up much faster, it stays at about 0.5 even after a long time of training. In fact, the maximum value dashed line in Figure~\ref{fig:return_comparison}(e) shows that none of the 10 runs of HER has reached a success rate of 0.6 or above. In contrast, although the empowerment approach is slower in the initial learning phase, in 3 out of 10 runs it has learned to lift up the object and reach the goals in the air accurately and quickly, and the success rate stays at about 1 in these tests. This is because in the Gym FetchPickAndPlace-V1 task, half of the goals are sampled from on the table and half are sampled in the air, thus agents that only learned to push can still reach the goals close to the tabletop and receive a success rate of about 0.5, but only agents that actually learned to pick and place will reach a success rate of 1.0.

\begin{figure}
\centering
 \includegraphics[width=0.9\linewidth]{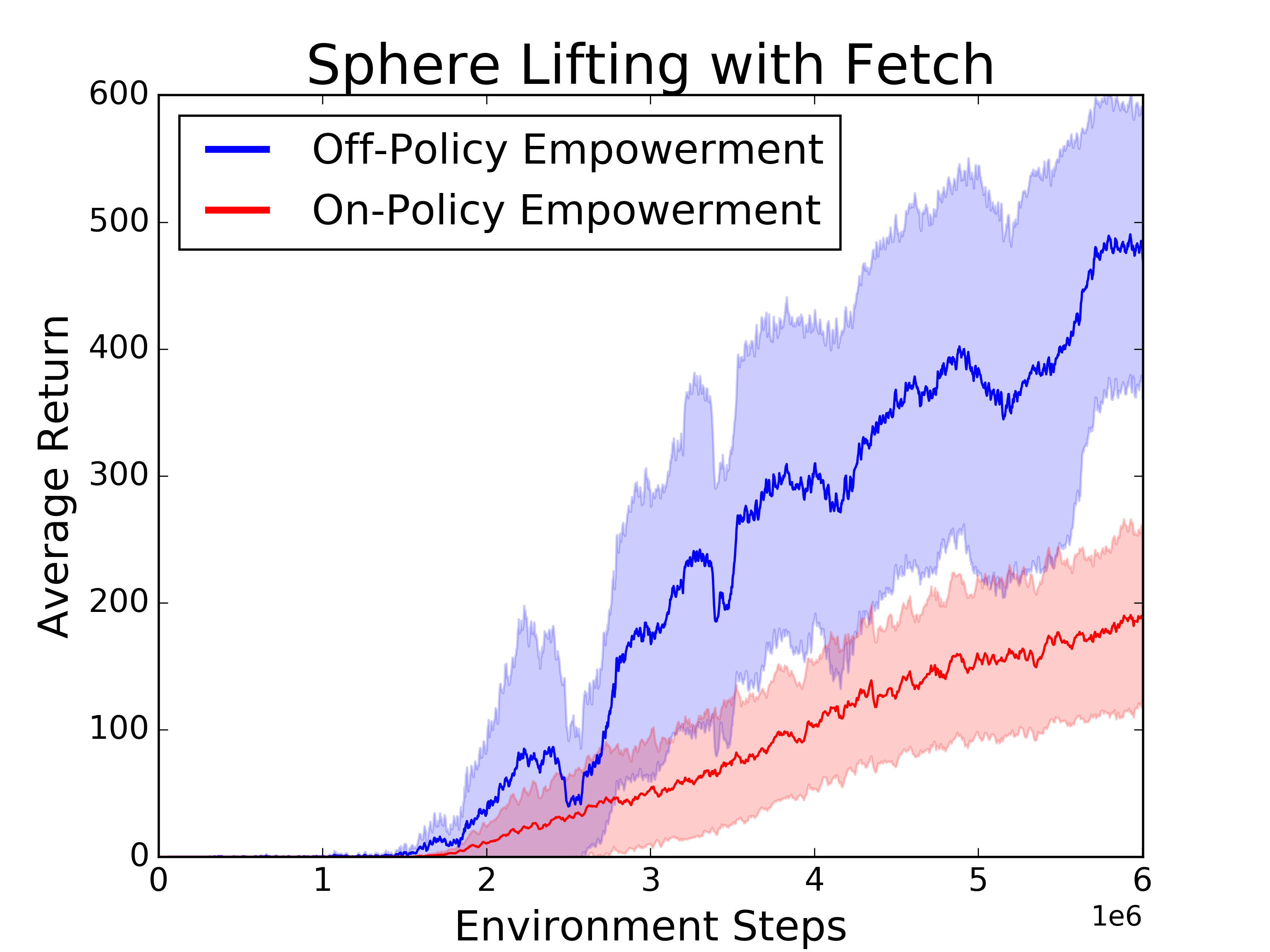}
 \caption{Comparison of off-policy implementation and on-policy implementation of the empowerment-based intrinsic exploration approach in the sphere lifting environment. The solid lines represent the mean of 10 experiments with different random seeds, and the shadow areas represent the 95\% confidence intervals.}
\label{fig:off_policy}
\end{figure}

% Figure~\ref{fig:return_comparison} compares the performance of empowerment-based intrinsic motivation with ICM~\citep{pathak2017curiosity}, exploration via disagreement~\citep{pathak2019self} (referred to as the Disagreement method), HER~\citep{andrychowicz2017hindsight}, and PPO without any intrinsic reward. 

   \begin{figure*}
      \centering
      \begin{subfigure}[t]{0.18\textwidth}
      \includegraphics[width=0.95\linewidth]{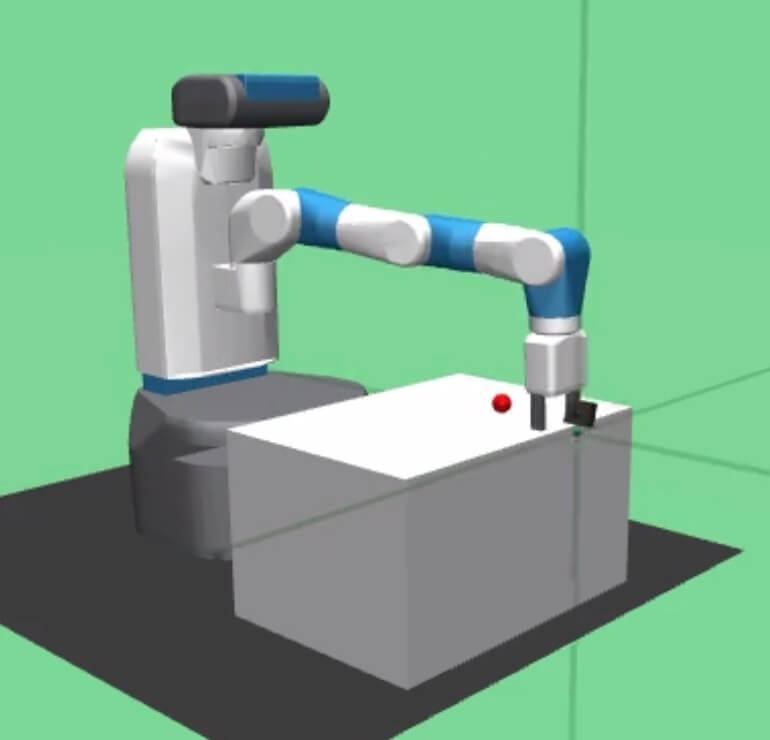}
      \caption{ }
      \end{subfigure}%
      \begin{subfigure}[t]{0.18\textwidth}
      \includegraphics[width=0.95\linewidth]{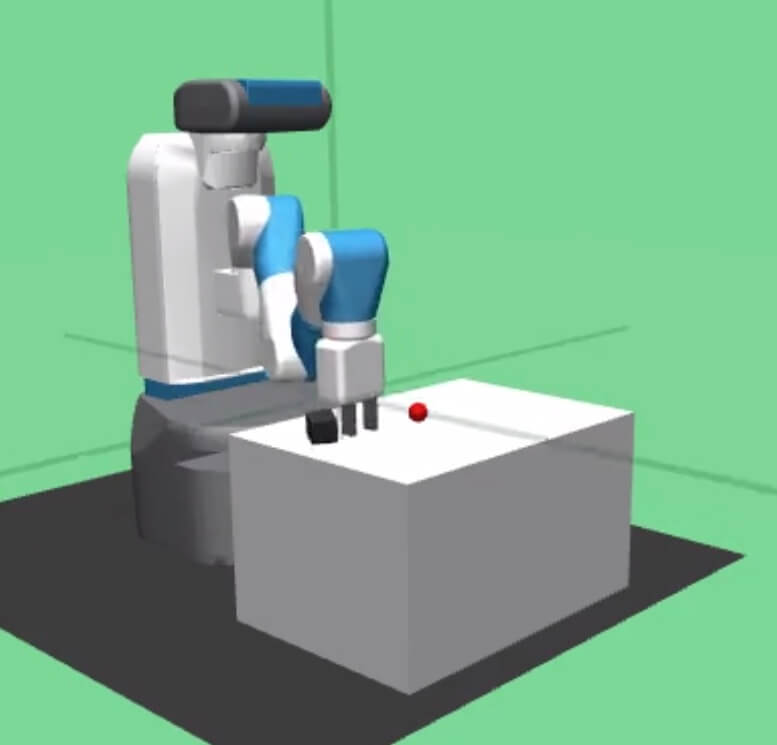}
      \caption{ }
      \end{subfigure}%
      \begin{subfigure}[t]{0.18\textwidth}
      \includegraphics[width=0.95\linewidth]{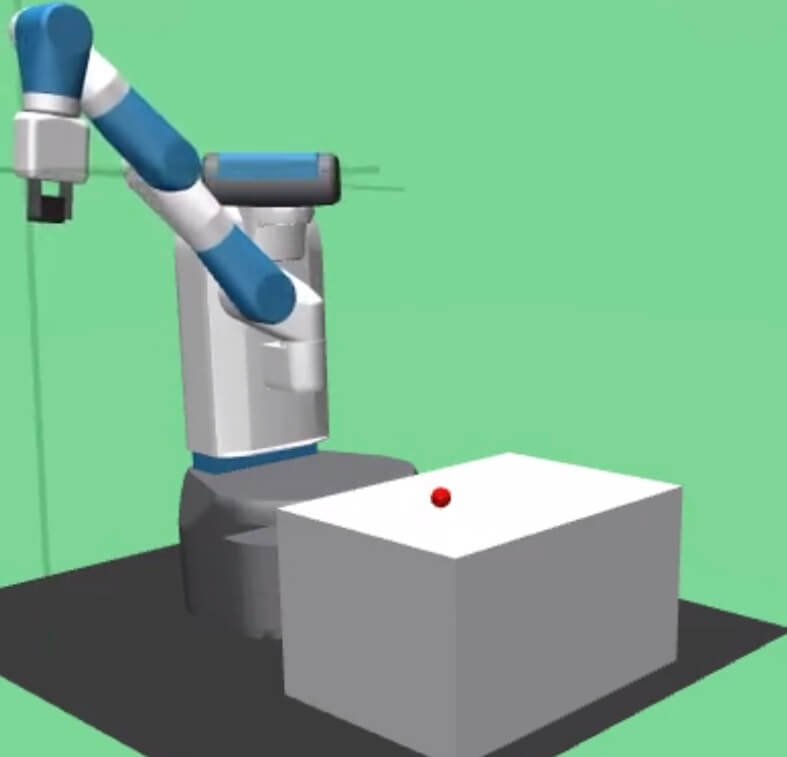}
      \caption{ }
      \end{subfigure}%
      \begin{subfigure}[t]{0.18\textwidth}
      \includegraphics[width=0.95\linewidth]{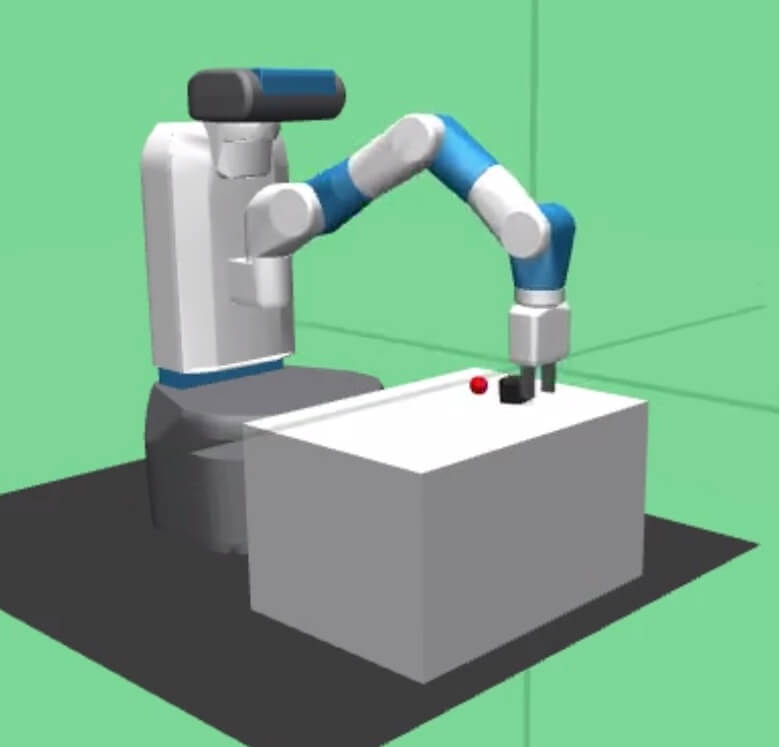}
      \caption{ }
      \end{subfigure}%
      \begin{subfigure}[t]{0.18\textwidth}
      \includegraphics[width=0.95\linewidth]{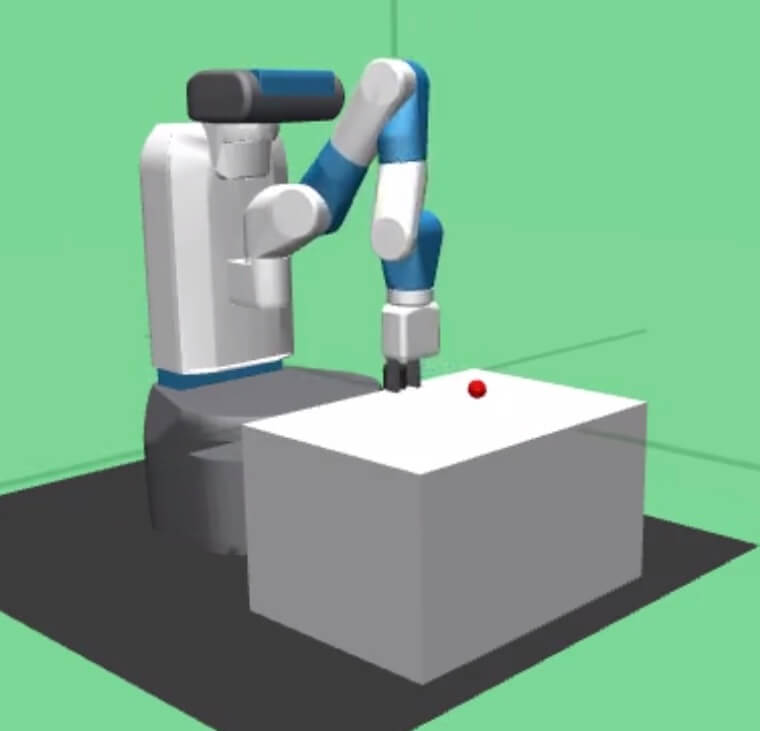}
      \caption{ }
      \end{subfigure}%
      \caption{Qualitative performance of the proposed empowerment-based intrinsic motivation when combined with the diversity-driven DIAYN~\citep{eysenbach2018diversity} approach in the box lifting task with a Fetch robot. (a)-(e) show the different skills learned when the number of skills in DIAYN is set to 5.}
      \label{fig:5skills}
   \end{figure*}

      \begin{figure*}
      \centering
      \begin{subfigure}[t]{0.18\textwidth}
      \includegraphics[width=0.95\linewidth]{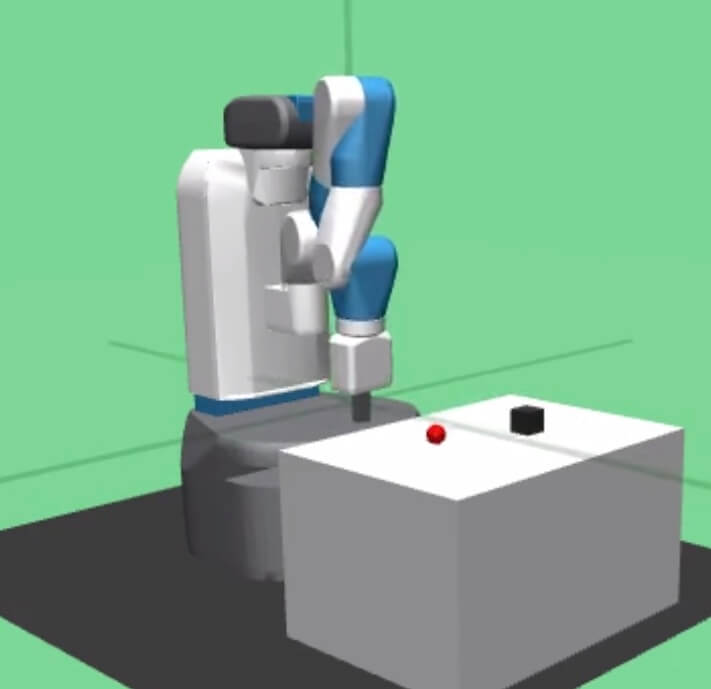}
      \caption{ }
      \end{subfigure}%
      \begin{subfigure}[t]{0.18\textwidth}
      \includegraphics[width=0.95\linewidth]{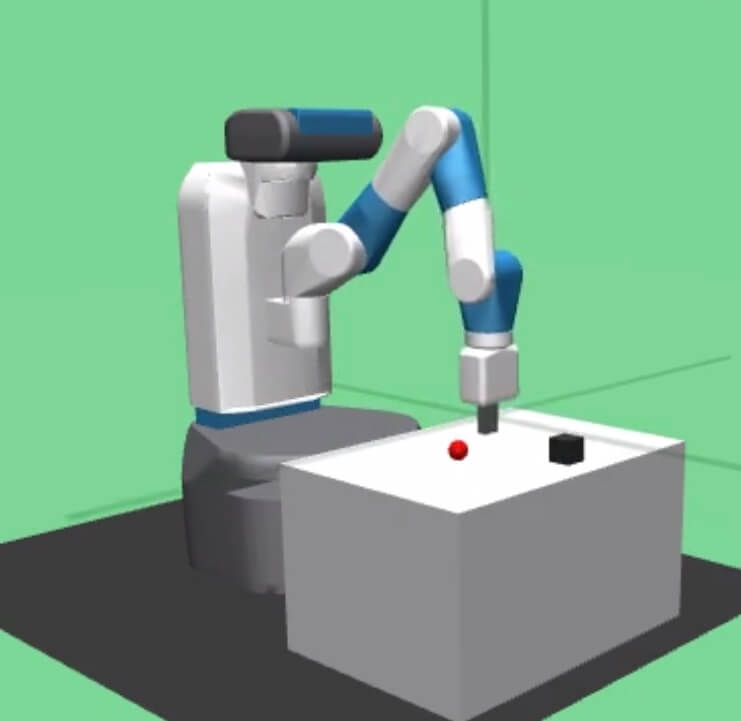}
      \caption{ }
      \end{subfigure}%
      \begin{subfigure}[t]{0.18\textwidth}
      \includegraphics[width=0.95\linewidth]{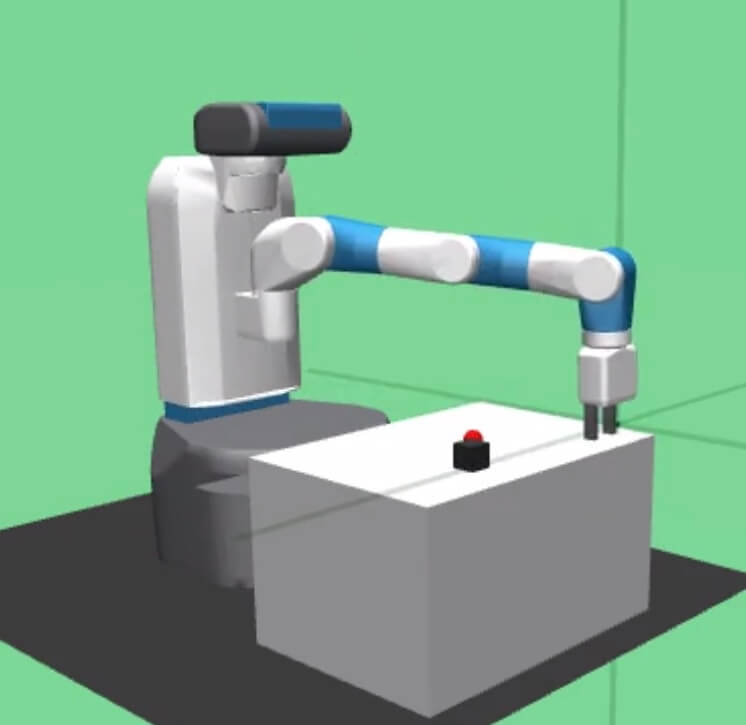}
      \caption{ }
      \end{subfigure}%
      \begin{subfigure}[t]{0.18\textwidth}
      \includegraphics[width=0.95\linewidth]{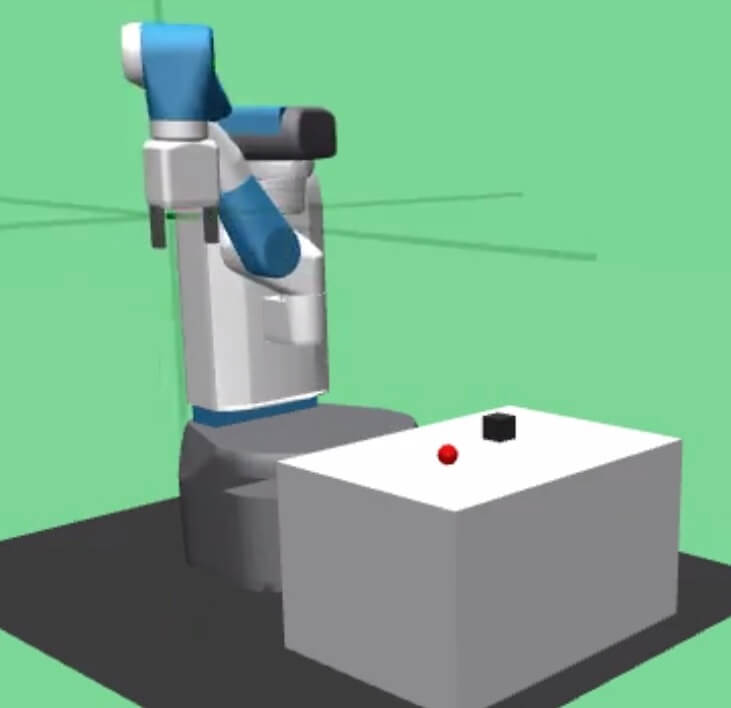}
      \caption{ }
      \end{subfigure}%
      \begin{subfigure}[t]{0.18\textwidth}
      \includegraphics[width=0.95\linewidth]{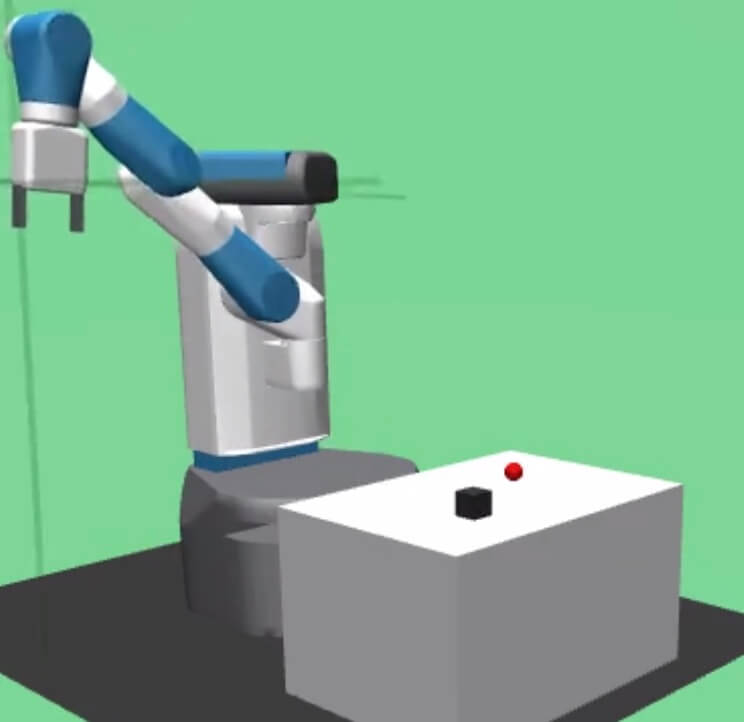}
      \caption{ }
      \end{subfigure}%
      \caption{Different skills learned with DIAYN~\citep{eysenbach2018diversity} without the empowerment-based intrinsic motivation in the box lifting task with a Fetch robot when the number of skills is set to 5.}
      \label{fig:diayn_5skills}
   \end{figure*}
   
\subsection{Off-Policy Implementation}

Our algorithm can also be used on off-policy RL algorithms but requires additional adaptation. This is because intrinsic rewards are not ``ground truth'' rewards and their values are not very meaningful until the neural networks are trained to predict intrinsic rewards well. Since the estimation of conditional mutual information is very challenging and the empowerment networks typically take a long time to get well trained, mixing up experiences with reward values predicted at different training steps in the same replay buffer will influence the overall performance and makes off-policy training very tricky. Therefore, we implemented an off-policy version by recomputing the intrinsic reward values with the updated network parameters every time the algorithm draws experiences from the replay buffer. We demonstrate its performance in the sphere lifting environment in Gym in Figure~\ref{fig:off_policy} and show that the off-policy implementation is much more sample-efficient.

\section{Application: Learning a Diverse Set of Skills}

% \subsection{Results with Diversity-Driven Intrinsic Motivation}

Besides its advantage in solving sparse reward RL tasks, another driving force for research on intrinsic motivation is its potential in unsupervised skill discovery. Many HRL frameworks allow RL agents to learn policies of different levels so that high-level policies only need to focus on the skill-space that low-level controllers provide instead of the raw state-space. However, the skills an end-to-end HRL system can learn are limited and they often require guidance from human-designed ``curricula''~\citep{bacon2017option, riedmiller2018learning, colas2019curious}. In contrast, skills discovered by intrinsic motivations can reduce HRL frameworks' dependence on human engineering and potentially enable them to learn more complicated tasks. Ultimately, we hope the empowerment-based intrinsic motivation proposed in this paper can also be incorporated into a HRL framework and contribute to the learning of complicated manipulation skills, such as opening a container and stacking objects inside.
In order to see what type of skills an agent can learn with our approach, we provide preliminary qualitative results combining empowerment and the Diversity is All You Need (DIAYN) approach~\citep{eysenbach2018diversity} in the ``Fetch with a box" environment. Figure~\ref{fig:5skills} and~\ref{fig:diayn_5skills} compare the skills learned by combining empowerment and DIAYN as the intrinsic reward and the skills learned with only DIAYN as the intrinsic reward. From Figure~\ref{fig:diayn_5skills} we can see that without an intrinsic motivation that drives the agent to control the object, the skills learned through a purely diversity-driven approach are not meaningful in terms of solving manipulation tasks because they don't involve interactions with the object. In comparison, Figure~\ref{fig:5skills} demonstrates the potential of this combined intrinsic reward in terms of learning a set of meaningful manipulation skills, including pushing the object to different directions and lifting the object up. Videos of the learned skills can be found at \url{https://sites.google.com/view/empowerment-for-manipulation/}.
   
% Besides accelerating the learning process of sparse reward tasks, one potential application for the empowerment-based intrinsic motivation proposed in this paper is to be learn a set of manipulation skills that can be transferred to new tasks via a hierarchical framework.  
%    
%    \begin{figure*}[h]
%       \centering
%       \begin{subfigure}[t]{0.2\textwidth}
%       \includegraphics[width=0.95\linewidth]{3skill_1.jpg}
%       \caption{ }
%       \end{subfigure}%
%       \begin{subfigure}[t]{0.2\textwidth}
%       \includegraphics[width=0.95\linewidth]{3skill_2.jpg}
%       \caption{ }
%       \end{subfigure}%
%       \begin{subfigure}[t]{0.2\textwidth}
%       \includegraphics[width=0.95\linewidth]{3skill_3.jpg}
%       \caption{ }
%       \end{subfigure}%
%       \caption{Combined Empowerment and DIAYN: 3 Skills}
%       \label{fig:3skills}
%    \end{figure*}

%===============================================================================

\section{Discussion}
% \label{sec:conclusion}

In this paper we present a novel intrinsic motivation for robotic manipulation tasks with sparse extrinsic rewards that leverages recent advances in both mutual information maximization and intrinsic novelty-driven exploration. Through maximizing the mutual dependence between robot actions and environment states, namely the empowerment, this intrinsic motivation helps the agent to focus more on the states where it can effectively ``control'' the environment instead of the parts where its actions cause random and unpredictable consequences. Despite the challenges posed by conditional mutual information maximization with continuous high-dimensional random variables, we are able to successfully train neural networks that make reasonable predictions on empowerment with the help of novelty-driven exploration methods at the beginning of the learning process. This approach can be easily integrated into any RL algorithm to accelerate their learning progress, or be combined with approaches like HER and imitation learning to further improve their performance.

Empirical evaluations in different robotic manipulation environments with different shapes of the target object demonstrate the advantages of this empowerment-based intrinsic motivation over other state-of-the-art solutions to sparse-reward RL tasks. In addition, we also combine this approach with diversity-driven intrinsic motivation and show that the combination is able to encourage the manipulator to learn a diverse set of ways to interact with the object, whereas with the diversity-driven rewards alone the manipulator is only able to learn how to move itself in different directions. In future work, we hope to apply this empowerment-based intrinsic motivation in a HRL framework that can utilize it to learn a diverse yet meaningful set of manipulation skills, so that the HRL agent can ultimately accomplish more complicated tasks that existing approaches can't learn from scratch without reward shaping or demonstrations, such as opening a container and stacking objects inside.

% \section*{Acknowledgments}

%% Use plainnat to work nicely with natbib. 

\bibliographystyle{plainnat}
\bibliography{references}

% \clearpage
% \appendix
% \appendixpage
\appendices
% \onecolumn

\section{Derivation of Mutual Information Lower Bounds} \label{sec:LowerBoundDerivation}

The VLB shown in Equation~\ref{equ:VariationalLowerBound} can be derived based on the non-negativity of KL-divergence:

\begin{equation}
\label{equ:VariationalLowerBoundDerivation}
 \begin{aligned}
  \mathcal{I}(X; Y) & = \mathbb{E}_{XY} \Big[ \log \frac{p(x|y) \cdot q(x|y)}{p(x) \cdot q(x|y)} \Big] \\
  & = \mathbb{E}_{XY} \Big[ \log \frac{q(x|y)}{p(x)} \Big] + \mathbb{E}_{XY} \Big[ \log \frac{p(x|y)}{q(x|y)} \Big] \\
  & = \mathbb{E}_{XY} \Big[ \log \frac{q(x|y)}{p(x)} \Big] + \mathbb{E}_Y \Big[ D_{KL} (p(x|y) || q(x|y)) \Big] \\
  & \geq \mathbb{E}_{XY} \Big[ \log \frac{q(x|y)}{p(x)} \Big]. \\
 \end{aligned}
\end{equation}

\noindent Similar to Equation~\ref{equ:Entropy}, conditional MI can also be written as:

\begin{equation}
\label{equ:ConditionalMIAlternatives}
 \begin{aligned}
  \mathcal{I}(X; Y|Z) & = \mathcal{H}(X|Z) - \mathcal{H}(X|Y, Z)\\
  & = \mathcal{H}(Y|Z) - \mathcal{H}(Y|X, Z) \\ 
  & = \mathcal{H}(X|Z) + \mathcal{H}(Y|Z) - \mathcal{H}(X, Y| Z) \\
  & = D_{KL}(p_{XY|Z}||p_{X|Z} p_{Y|Z}), \\
 \end{aligned}
\end{equation}

\noindent For conditional MI $\mathcal{I}(X; Y| Z)$, the VLB can be derived as:

\begin{equation}
\label{equ:ConditionalVariationalLowerBound}
% \small
 \begin{aligned}
  \mathcal{I}(X; Y|Z) = & \mathbb{E}_{XY|Z} \Big[ \log \frac{p(x|y, z) \cdot q(x|y, z)}{p(x|z) \cdot q(x|y, z)} \Big] \\
  = & \mathbb{E}_{XY|Z} \Big[ \log \frac{q(x|y, z)}{p(x|z)} \Big] + \mathbb{E}_{XY|Z} \Big[ \log \frac{p(x|y, z)}{q(x|y, z)} \Big] \\
  = & \mathbb{E}_{XY|Z} \Big[ \log \frac{q(x|y, z)}{p(x|z)} \Big] \\
  & + \mathbb{E}_{Y|Z} \Big[ D_{KL} (p(x|y, z) || q(x|y, z)) \Big] \\
  \geq & \mathbb{E}_{XY|Z} \Big[ \log \frac{q(x|y, z)}{p(x|z)} \Big]. \\
 \end{aligned}
\end{equation}

\noindent where $q(x|y, z)$ is a variational approximation of $p(x|y, z)$, and the bound is tight when $q(x|y, z) = p(x|y, z)$. 

% Other variational lower bounds of MI have also been derived based on a broader class of distance measures called $f$-divergence~\cite{liese2006divergences, nguyen2010estimating, nowozin2016f}. 
% The variational lower bound of $f$-divergences has been derived in \cite{} and \cite{}:
% \begin{equation}
% \label{equ:f_bound}
%  D_{f}(P(z)||Q(z)) \geq \sup_{T \in \mathcal{T}} (\mathbb{E}_{z \sim P}[T(z)] - \mathbb{E}_{z \sim Q}[f^* (T(z))]),
% \end{equation}
% \noindent where $\mathcal{T}$ is and arbitrary class of functions $T: \mathcal{Z} \rightarrow \mathbb{R}$, and $f^*$ is the convex conjugate of $f$. KL-divergence is a special case of $f$-divergence when the generator function $f(u) = u \log u$, and Jensen-Shannon (JS) divergence is a special case when $f(u)=-(u+1)\log((1+u)/2) + u\log u$~\cite{nowozin2016f}.
The $f$-divergence between two distributions $P$ and $Q$ is defined as:

\begin{equation}
 D_{f}(P(z)||Q(z)) = \int f \bigg( \frac{dP}{dQ} \bigg) \,dQ  = \int_z f \bigg( \frac{p(z)}{q(z)} \bigg) q(z) \,dz,
\end{equation}

\noindent where the generator function $f: \mathbb{R}_+ \rightarrow \mathbb{R}$ is a convex, lower-semicontinuous function satisfying $f(1) = 0$. The variational lower bound of $f$-divergences has been derived in \cite{nguyen2010estimating} and \cite{nowozin2016f}:

\begin{equation}
\label{equ:f_bound}
 D_{f}(P(z)||Q(z)) \geq \sup_{T \in \mathcal{T}} (\mathbb{E}_{z \sim P}[T(z)] - \mathbb{E}_{z \sim Q}[f^* (T(z))]),
\end{equation}

\noindent where $\mathcal{T}$ is and arbitrary class of functions $T: \mathcal{Z} \rightarrow \mathbb{R}$, and $f^*$ is the convex conjugate of $f$. Equation~\ref{equ:f_bound} yields a lower bound because the class of functions $\mathcal{T}$ may only contain a subset of all possible functions, and under mild conditions on $f$~\cite{nguyen2010estimating}, the bound is tight when: 

\begin{equation}
\label{equ:TBound}
 T(x) = f' \bigg( \frac{p(z)}{q(z)} \bigg).
\end{equation}

% KL-divergence and Jensen-Shannon (JS) divergence are two special cases of $f$-divergence.
% Based on the relationship between MI and KL-divergence shown in Equation~\ref{equ:Entropy}, a lower bound of MI is derived in~\cite{belghazi2018mutual}:
KL-divergence is a special case of $f$-divergence when the generator function $f(u) = u \log u$~\cite{nowozin2016f}. Therefore, a lower bound of KL-divergence can be derived as Equation~\ref{equ:KLD}. For conditional MI $\mathcal{I}(X; Y| Z)$, the KLD lower bound can be written as:

\begin{equation}
\label{equ:ConditionalKLD}
 \mathcal{I}_{KL}(X; Y|Z) \geq \sup_{T \in \mathcal{T}} \mathbb{E}_{p_{XY|Z}} [T] - \mathbb{E}_{p_{X|Z} p_{Y|Z}}[e^{T - 1}],
\end{equation}

\noindent where $\mathcal{T}$ is an arbitrary class of functions $T: \mathcal{X} \times \mathcal{Y} \times \mathcal{Z} \rightarrow \mathbb{R}$.
% \begin{equation}
% \label{equ:KLDDerivation}
%  \mathcal{I}_{KL}(X; Y) \geq \sup_{T \in \mathcal{T}} \mathbb{E}_{p_{XY}} [T] - \mathbb{E}_{p_{X} p_{Y}}[e^{T - 1}],
% \end{equation}

% \noindent where $\mathcal{T}$ is an arbitrary class of functions $T: \mathcal{X} \times \mathcal{Y} \rightarrow \mathbb{R}$. 

Jensen-Shannon (JS) divergence is another special case of $f$-divergence. It can be expressed in terms of KL-divergence:

\begin{equation}
 D_{JS}(P||Q) = \frac{1}{2} D_{KL} (P||M) + \frac{1}{2} D_{KL} (Q||M),
\end{equation}

\noindent where $M=1/2(P+Q)$. JS-divergence represents the mutual information between a random variable $A$ associated to a mixture distribution between $P$ and $Q$ and a binary indicator variable $B$ that is used to switch between $P$ and $Q$. In particular, if we use $P$ to represent the joint distribution $p_{XY}$ and use $Q$ to represent the product of the marginal distributions $p_X p_Y$, then:

\begin{equation}
\begin{aligned}
 p(A|B) = & \begin{cases} p(x, y) & \textrm{if} ~~ B = 0, \\
           p(x)p(y) & \textrm{if} ~~ B = 1. \\
          \end{cases}
\end{aligned}
\end{equation}

\noindent That is, the random variable $A$ is chosen according to the probability measure $M = (P+Q)/2$, and its distribution is the mixture distribution. Then the relationship between JS-divergence and mutual information can be derived as follows:

\begin{equation}
\label{equ:JS}
\begin{aligned}
 \mathcal{I}(A; B) = & \mathcal{H}(A) - \mathcal{H}(A|B) \\
 = & - \sum M \log M + \frac{1}{2} [\sum P \log P + \sum Q \log Q ] \\
 = & - \sum \frac{P}{2} \log M - \sum \frac{Q}{2} \log M \\
 & + \frac{1}{2} [\sum P \log P + \sum Q \log Q ] \\
 = & \frac{1}{2} \sum P(\log P - \log M) \\
 & + \frac{1}{2} \sum Q(\log Q - \log M) \\
 = & D_{JS}(P||Q). \\
\end{aligned}
\end{equation}

\noindent Therefore, if we define the Jensen-Shannon mutual information (JSMI) between two random variables $X$ and $Y$ as the JS-divergence between their joint distribution and the product of their marginal distributions, i.e. $\mathcal{I}_{JS}(X; Y) \equiv D_{JS} (p_{XY}||p_X p_Y)$, then Equation~\ref{equ:JS} shows that:

\begin{equation}
 \mathcal{I}_{JS}(X; Y) = \mathcal{I}(A; B).
\end{equation}

\begin{table*}[h]
\caption{Comparison of Mutual Information Lower Bounds}
\label{table:MI_Unit_Test_Results}
 \centering
 \begin{tabular}{L{1.5cm}|L{2cm}|L{1.5cm}|L{1.5cm} L{1.5cm} L{1.5cm}}
 \toprule
 \multirow{2}{1.5cm}{\centering Dimension} & \multirow{2}{2cm}{\centering Theoretical Average MI} & \multirow{2}{1.5cm}{\centering Training Data Size} & \multicolumn{3}{c}{Root Mean Square Error (RMSE)} \\ \cline{4-6}
 & & & VLB & KLD & JSD \\ \hline
 \multirow{4}{1.5cm}{\centering 1} & \multirow{4}{2cm}{\centering 0.2911} & 20000 & 0.0713 & 0.1661 & 0.1594 \\ 
 & & 40000 & 0.0424 & 0.1291 & 0.1242 \\ 
 & & 60000 & 0.0502 & 0.1509 & 0.1785 \\ \hline
 \multirow{4}{1.5cm}{\centering 2} & \multirow{4}{2cm}{\centering 0.5821} & 20000 & 0.0974 & 0.3745 & 0.2578 \\ 
 & & 40000 & 0.1121 & 0.3517 & 0.3292 \\ 
 & & 60000 & 0.0942 & 0.2139 & 0.2105 \\ \hline
 \multirow{4}{1.5cm}{\centering 3} & \multirow{4}{2cm}{\centering 0.8732} & 20000 & 0.1594 & 0.4825 & 0.4573 \\ 
 & & 40000 & 0.1508 & 0.4828 & 0.4573 \\ 
 & & 60000 & 0.1407 & 0.4129 & 0.3176 \\ \hline
 \multirow{4}{1.5cm}{\centering 4} & \multirow{4}{2cm}{\centering 1.1643} & 20000 & 0.2222 & 0.5879 & 0.5406 \\ 
 & & 40000 & 0.1665 & 0.6092 & 0.4101 \\ 
 & & 60000 & 0.1611 & 0.4928 & 0.4326 \\ 
 \bottomrule
 \end{tabular}
\end{table*}

The advantage of using JS-divergence is that it is not only symmetric but also bounded from both below and above~\cite{kim2019emi}. Although different from the commonly accepted definition of MI, JSMI is closely correlated to MI and can also represent the mutual dependence between random variables. It is shown in \cite{nowozin2016f} that JS-divergence is a special case of $f$-divergence when the generator function $f(u)=-(u+1)\log((1+u)/2) + u\log u$, hence its lower bound can be derived as:

% \begin{equation}
% \label{equ:JSD}
% \begin{aligned}
%  \mathcal{I}_{JS}(X; Y) & = D_{JS} (p_{XY}||p_X p_Y) \\
%  & \geq \sup_{T \in \mathcal{T}} \mathbb{E}_{p_{XY}} [\log 2 - \log (1 + e^{-T})] - \mathbb{E}_{p_X p_Y} [D_{JS}^* (\log 2 - \log (1 + e^{-T}))] \\
%  & = \sup_{T \in \mathcal{T}} \mathbb{E}_{p_{XY}} [-\textrm{sp} (-T)] - \mathbb{E}_{p_X p_Y} [\textrm{sp} (T)] + \log 4, \\
% \end{aligned}
% \end{equation}
% 
% \noindent where $D_{JS}^* (u) = -\log (2-\exp(u))$ is the Fenchel conjugate of JS-divergence, and $\textrm{sp} (u) = \log (1+\exp (u))$ is the soft plus function.

% The JS definition of MI is closely related to the MI we defined in Equation~\ref{equ:MI_def}, and its lower bound can be derived as~\cite{kim2019emi}:

\begin{equation}
\label{equ:JSDDerivation}
\begin{aligned}
 \mathcal{I}_{JS}(X; Y) = & D_{JS} (p_{XY}||p_X p_Y) \\
 \geq & \sup_{T \in \mathcal{T}} \mathbb{E}_{p_{XY}} [\log 2 - \log (1 + e^{-T})] \\
 & - \mathbb{E}_{p_X p_Y} [D_{JS}^* (\log 2 - \log (1 + e^{-T}))] \\
 = & \sup_{T \in \mathcal{T}} \mathbb{E}_{p_{XY}} [-\textrm{sp} (-T)] - \mathbb{E}_{p_X p_Y} [\textrm{sp} (T)] + \log 4, \\
\end{aligned}
\end{equation}

\noindent where $D_{JS}^* (u) = -\log (2-\exp(u))$ is the Fenchel conjugate of JS-divergence, and $\textrm{sp} (u) = \log (1+\exp (u))$ is the soft plus function. 
% Note that Equation~\ref{equ:JSDDerivation} is not a lower bound for the MI we defined in Equation~\ref{equ:MI_def}, but since the two MIs are closely related, it is also often used to estimate the MI defined in Equation~\ref{equ:MI_def}. 
% In this paper, we refer to the variational lower bound in Equation~\ref{equ:VariationalLowerBoundDerivation} as VLB, the lower bound based on KL-divergence in Equation~\ref{equ:KLDDerivation} as KLD, and the lower bound for JS-divergence based mutual information in Equation~\ref{equ:JSDDerivation} as JSD. 
The JSD lower bound for conditional MI can be written as:

\begin{equation}
\label{equ:ConditionalJSD}
\begin{aligned}
 \mathcal{I}_{JS}(X; Y|Z) = & D_{JS} (p_{XY|Z}||p_{X|Z} p_{Y|Z}) \\
 \geq & \sup_{T \in \mathcal{T}} \mathbb{E}_{p_{XY|Z}} [\log 2 - \log (1 + e^{-T})] \\
 & - \mathbb{E}_{p_{X|Z} p_{Y|Z}} [D_{JS}^* (\log 2 - \log (1 + e^{-T}))] \\
 = & \sup_{T \in \mathcal{T}} \mathbb{E}_{p_{XY|Z}} [-\textrm{sp} (-T)] \\
 & - \mathbb{E}_{p_{X|Z} p_{Y|Z}} [\textrm{sp} (T)] + \log 4, \\
\end{aligned}
\end{equation}

\noindent where $\mathcal{T}$ is an arbitrary class of functions $T: \mathcal{X} \times \mathcal{Y} \times \mathcal{Z} \rightarrow \mathbb{R}$. Following Equation~\ref{equ:TBound} we can then derive that the bound for conditional JSD is tight when:

\begin{equation}
 T(x) = f' \bigg( \frac{p(x, y|z)}{p(x|z)p(y|z)} \bigg), 
\end{equation}

\noindent hence $T$ can be used as the empowerment intrinsic reward if we maximize the conditional JSD bound in Equation~\ref{equ:ConditionalJSD}.

\section{Comparison of Mutual Information Lower Bounds}
\label{sec:LowerBoundComparison}

We construct a set of distributions with known theoretical MI:

\begin{equation}
 \begin{aligned}
 Z \sim & ~ \mathcal{N}(0, \sigma_z^2), ~ X = ~ Z + e, ~ e \sim ~ \mathcal{N}(0, 1), \\
 Y = & \begin{cases} Z + X \cdot Z + f & \textrm{if} ~ Z > 0, \\
        f & \textrm{if} ~ Z \leq 0, \\
       \end{cases} 
 ~ f \sim ~ \mathcal{N}(0, n^2).  \\
 \end{aligned}
\end{equation}

\noindent Based on the theoretical MI for bivariate Gaussian distributions~\cite{gel1959abouta}, we can compute the conditional MI:

\begin{equation}
\label{equ:TheoreticalMI}
 \mathcal{I}(X; Y|Z)  = \frac{1}{2} \log (1 + \frac{z^2}{n^2}).
\end{equation}

\noindent We conduct tests on the $X$, $Y$ and $Z$ random variables described above with $\sigma_z = 1$ and $n = 0.5$. We compare the performance of the three different estimation approaches introduced in Section~\ref{sec:MI} given different variable dimensions and different sizes of training data, and evaluate them using the root mean square error (RMSE) compared to the theoretical value of MI computed through Equation~\ref{equ:TheoreticalMI}.
We use a neural network with one hidden layer of 256 units as the MI estimator for each approach. We compare the performance of the three different estimation approaches given different variable dimensions and different sizes of training data, and the results are shown in Table~\ref{table:MI_Unit_Test_Results}. The performance of each estimation approach is evaluated based on the root mean square error (RMSE) compared to the theoretical value of MI computed through Equation~\ref{equ:TheoreticalMI}. 

From Table~\ref{table:MI_Unit_Test_Results} we can see that the VLB has the lowest RMSE in all the test cases on this random variable set, whereas the KLD bound performs the worst in most cases. From the comparison between the RMSE and the absolute values of theoretical average MI we can see that it is possible to get a relatively accurate approximation of the conditional MI through numerical estimation when the mutual dependency between random variables are simple. 
% In the robotics manipulation experiments in this paper, we noticed that JSD is the best performer on Fetch robot and VLB is the best performer on PR2, hence we will only report the results with the corresponding best performer in each experiment.
% It also doesn't prove that Variational bound will provide the best estimations for all distributions, but it did indicate the KLD bound can't provide a stable and relatively accurate estimation even for simple distributions. Therefore, in the robotics experiments in this thesis, we will consider the Variational bound and the JSD bound only for mutual information estimation.

% As mentioned in Section~\ref{sec:MI}, MI estimation is notoriously difficult for continuous random variables, and empowerment estimation is even more challenging because it requires the computation of MI conditioned on another continuous random variable. Since in most reinforcement learning scenarios, we typically don't have the exact distribution to calculate MI and numerical estimation through sampling is required, conditioning on a continuous random variable means that we almost always only have one sample per pair of distributions to estimate the conditional MI. 

\section{Experiment Details}
\label{sec:details}

For the experiments shown in this paper, we implemented the empowerment-based approach, the ICM approach and the Disagreement approach as intrinsic rewards with an on-policy implementation of PPO. 
% We use on-policy PPO because intrinsic rewards are not ``ground truth'' rewards and their values are not very meaningful until the neural networks are trained to predict intrinsic rewards well. Since the estimation of conditional mutual information is very challenging and the empowerment networks typically take a long time to get well trained, mixing up experiences with reward values predicted at different training steps in the same replay buffer will influence the overall performance and makes off-policy training very tricky. 
We use a three hidden-layer fully-connect neural network with (128, 64, 32) units in each layer for both the policy network and the value network, and set $\gamma = 0.99$ and $\lambda = 0.95$ in the PPO algorithm. We use the Adam optimizer with learning rate $2\mathrm{e}{-4}$. All experiments shown in this paper are conducted on a 10-core Intel i7 3.0 GHz desktop with 64 GB RAM and one GeForce GTX 1080 GPU.

\paragraph{ICM Implementation}
In the experiments in this paper, since we assume pose estimations are available, the inverse model of ICM is not necessary. In the ICM implementation, we train the forward model $f$ by minimizing the forward loss:

\begin{equation}
\label{equ:forward_loss}
 \mathcal{L}_{t}^f = \frac{1}{2} || f \big( \mathbf{s}_t^{ex}, \mathbf{a}_t \big) - \mathbf{s}_{t+1}^{ex} ||^2_2. 
\end{equation}

\noindent To compute the forward loss in the ICM approach, we use one 256-unit hidden layer in the network, and we didn't compute inverse loss because the observations in this paper are poses instead of images. The value of the forward loss $\mathcal{L}_{t}^f$ is also used as the ICM intrinsic reward:

\begin{equation}
\label{equ:r_ICM}
 r_t^{ICM} = \mathcal{L}_{t}^f,
\end{equation}

\noindent and we normalize $r_t^{ICM}$ using running average before summing it up with the extrinsic reward to get the final reward for training the RL agent:

\begin{equation}
 r_t = 0.01 \bar{r}_t^{ICM} + r_t^e.
\end{equation}

\paragraph{Disagreement Implementation}
In the Disagreement approach, we use the same network structure as in ICM and use five of these networks as the ensemble to compute the disagreement reward. We compute the forward losses for each of the five forward models in the same way as Equation~\ref{equ:forward_loss}, and sum up the five forward losses as the total loss to train the forward models. The intrinsic reward is calculated as:

\begin{equation}
 r_t^{Dis} = var \{ \hat{\mathbf{s}}_{t+1}^{ex,1}, \dots, \hat{\mathbf{s}}_{t+1}^{ex,5} \},
\end{equation}

\noindent where $\hat{\mathbf{s}}_{t+1}^{ex,1}$ through $\hat{\mathbf{s}}_{t+1}^{ex,5}$ are the forward predictions made by the five forward models. We also use running average to get the normalized disagreement intrinsic reward $\bar{r}_t^{Dis}$ and then sum it up with the extrinsic reward to get the final reward for training the RL agent:

\begin{equation}
 r_t = 0.01 \bar{r}_t^{Dis} + r_t^e.
\end{equation}

\paragraph{Empowerment Implementation}
For the neural network that makes empowerment prediction in the PR2 environment, we apply Gated Linear Units (GLU)~\citep{dauphin2017language} to improve performance. We use a neural network with four GLU layers with 256 gates each and two hidden fully-connected layers with (128, 64) units to predict $p(\mathbf{a}_t | \mathbf{s}^{ex}_{t+1}, \mathbf{s}_t)$, and calculate empowerment with the variational lower bound. Namely, we use

\begin{equation}
 r_t^{Emp} = \log p(\mathbf{a}_t | \mathbf{s}^{ex}_{t+1}, \mathbf{s}_t) - \log p(\mathbf{a}_t | \mathbf{s}_t)
\end{equation}

\noindent as the empowerment intrinsic reward so that in expectation, the empowerment reward being maximized is equivalent to the empowerment defined in Equation~\ref{equ:approx_empowerment}. In the Fetch environment, we use a neural network with six hidden fully-connected layers with (512, 512, 216, 128, 64, 32) units to approximate the $T$ function in Equation~\ref{equ:ConditionalJSD} and calculate empowerment with the JS-Divergence approximation. In order to approximate the supremum in Equation~\ref{equ:ConditionalJSD}, we use the following loss function in order to train $T$ network:

\begin{equation}
 \mathcal{L}_t^{Emp} = \textrm{sp} (- T(\mathbf{a}_t, \mathbf{s}_t, \mathbf{s}^{ex}_{t+1})) + \textrm{sp} (T(\tilde{\mathbf{a}}_t, \mathbf{s}_t, \mathbf{s}^{ex}_{t+1})) - \log 4,
\end{equation}

\noindent where $\mathbf{a}_t$ is the true action executed at time step $t$ and $\tilde{\mathbf{a}}_t$ is sampled from the policy. The empowerment intrinsic reward in the Fetch environment is:

\begin{equation}
 r_t^{Emp} = T(\mathbf{a}_t, \mathbf{s}_t, \mathbf{s}^{ex}_{t+1}).
\end{equation}

In our empowerment-based intrinsic motivation implementation, empowerment reward and ICM reward are combined through weight coefficients to ensure that the agent can collect enough data in the nonzero empowerment region to train the empowerment network well before it is used as the intrinsic reward. The weight coefficients used in this paper are:

\begin{equation}
\begin{aligned}
 & w_t^{ICM} = 0.5 \times (1 - \tanh (200 (r_t^{ICM} - 0.12))), \\
 & w_t^{Emp} = 1 -  w_t^{ICM}, \\
\end{aligned}
\end{equation}

\noindent where $r_t^{ICM}$ is the forward prediction error (computed through Equation~\ref{equ:forward_loss} and \ref{equ:r_ICM}) averaged from all the parallel environments at time step $t$. These weight coefficients make sure that at the beginning of training when the robot don't have much interaction with the object, the coefficient for ICM reward is near 1 and the coefficient for empowerment reward is near 0. After the average ICM reward reaches a certain threshold, which means the robot have learned to interact with the object and the empowerment network can obtain enough meaningful data to get well trained, the coefficient for ICM reward switches to near 0 and the coefficient of the empowerment reward switches to near 1. Then this intrinsic reward and extrinsic task reward are combined as the RL algorithm reward:

\begin{equation}
 \begin{aligned}
  r_t^i & = w_t^{ICM} \bar{r}_t^{ICM} + w_t^{Emp} \bar{r}_t^{Emp}, \\
  r_t & = 0.01 r_t^i + r_t^e, \\
 \end{aligned}
\end{equation}

\noindent where $\bar{r}_t^{ICM}$ and $\bar{r}_t^{Emp}$ are normalized using running average.

\paragraph{Extrinsic Task Rewards}
In the box-lifting task and the pick-and-place task in the Fetch environment, the object is a cube with 0.05 m edges. In the cylinder-lifting environment, the height of the cylinder is 0.1 m and the radius is 0.03 m. In the sphere-lifting environment, the radius of the sphere is 0.04 m. In both the box-lifting and sphere-lifting task, the task reward is given as Equation~\ref{equ:Fetch_reward} when the center of the grippers is less than 0.01 m away from the center of the object. In the cylinder-lifting task, the condition for giving task reward is the same, but the reward is given as Equation~\ref{equ:cylinder_reward}. In the pick-and-place task, the task reward is 1 when the object pose is within 0.05 m of the target pose, and 0 otherwise.

\begin{equation}
\label{equ:Fetch_reward}
 \textrm{Fetch with box or sphere:}~~ r_t^e = 50 \cdot (h - 0.01),
\end{equation}

\begin{equation}
\label{equ:cylinder_reward}
 \textrm{Fetch with cylinder:}~~ r_t^e = 500 \cdot (h - 0.01),
\end{equation}

In the box-lifting task in the PR2 environment, the object is a cube with 0.06 m edges, and the task reward is given as Equation~\ref{equ:PR2_reward} when both grippers are in contact with the object and the object height is at least 0.012 m above the tabletop.

\begin{equation}
\label{equ:PR2_reward}
 \textrm{PR2 with box:}~~ r_t^e = 500 \cdot (h - 0.012).
\end{equation}

\end{document}